\documentclass[journal,full]{IEEEtran}
\IEEEoverridecommandlockouts  
\usepackage{graphics} 
\usepackage[line
snumbered,ruled]{algorithm2e}
\graphicspath{{img/}}
\usepackage{epsfig} 
\usepackage{subfigure}
\usepackage{amsmath} 
\usepackage{amssymb}  
\usepackage{amsthm}
\usepackage{bigints}
\usepackage{pifont}
\def\-{\raisebox{.75pt}{-}}
\usepackage{mathtools}
\usepackage{cite}
\usepackage[makeroom]{cancel}

\usepackage{multirow}
\usepackage{tabularx}
\usepackage{makecell}
\usepackage{units}
\usepackage[table]{xcolor}
\usepackage{color}
\definecolor{Gray}{gray}{0.95}
\usepackage{hyperref}
\usepackage{booktabs}
\usepackage{epstopdf}
\usepackage{textcomp}
\usepackage{bm}
\usepackage{caption}

\usepackage{algpseudocode,algorithm} 
\usepackage{siunitx} 
\usepackage{mathptmx}
\usepackage[11pt]{moresize}
\usepackage{hhline}
\usepackage{tikz}
\usepackage{textcomp}
\def\BibTeX{{\rm B\kern-.05em{\sc i\kern-.025em b}\kern-.08em
    T\kern-.1667em\lower.7ex\hbox{E}\kern-.125emX}}

\newcommand{\etal}{\textit{et al}.}

\let\oldtwocolumn\twocolumn
\renewcommand\twocolumn[1][]{%
	\oldtwocolumn[{#1}{
		\begin{center}
			\includegraphics[width=\textwidth]{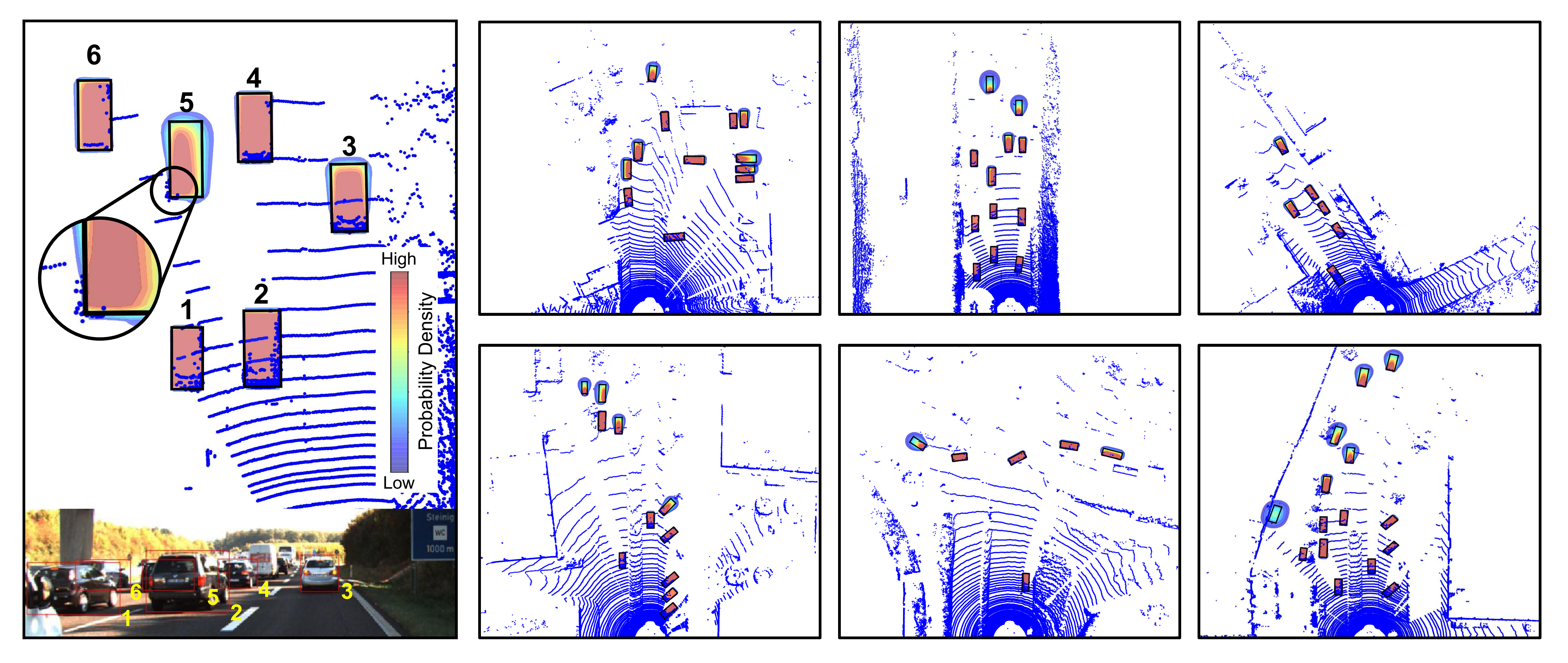}
			\captionof{figure}{Illustration of our proposed label uncertainty in bounding boxes in the KITTI dataset~\cite{geiger2012cvpr}. Original ground truth bounding boxes (black) are shown on the LiDAR Bird's Eye View (BEV). They are deterministic and do not provide information on their quality or environmental noise. However, there exist errors (or uncertainty) inherent in labels. In the leftmost figure, determining the length of object $3$ is difficult, because there are LiDAR reflections only on the side facing towards the ego-vehicle; Annotating object $5$ is more challenging, since the LiDAR reflections only cover its bottom-left corer. In this work, we build a generative model of LiDAR points to infer label uncertainty, shown by the spatial uncertainty distribution around each bounding box.}\label{fig:introduction}
		\end{center}
	}]
}

\begin{document}
\title{\Huge Labels Are Not Perfect: Inferring Spatial Uncertainty in Object Detection
\thanks{$^{\ast}$ Di Feng and Zining Wang contributed equally to this work.}
\thanks{$^1$ Mechanical Systems Control Lab, University of California, Berkeley, CA, 94720, USA.}
\thanks{$^2$ Institute of Measurement, Control and Microtechnology, Ulm University, 89081, Ulm, Germany.}
\thanks{$^3$ Robert Bosch GmbH, Corporate Research, Driver Assistance Systems and Automated Driving, 71272, Renningen, Germany.}
\thanks{$^{\ast}$ Equal contribution. Correspondence: \texttt{di.feng@uni-ulm.de}.}
}
\author{Di Feng$^{\ast1,2}$, Zining Wang$^{\ast1}$, Yiyang Zhou$^{1}$, Lars Rosenbaum$^{3}$, Fabian Timm$^{3}$, \\Klaus Dietmayer$^{2}$, Masayoshi Tomizuka$^{1}$, and Wei Zhan$^{1}$}

\maketitle

\begin{abstract}
The availability of many real-world driving datasets is a key reason behind the recent progress of object detection algorithms in autonomous driving. However, there exist ambiguity or even failures in object labels due to error-prone annotation process or sensor observation noise. Current public object detection datasets only provide deterministic object labels without considering their inherent uncertainty, as does the common training process or evaluation metrics for object detectors. As a result, an in-depth evaluation among different object detection methods remains challenging, and the training process of object detectors is sub-optimal, especially in probabilistic object detection. In this work, we infer the uncertainty in bounding box labels from LiDAR point clouds based on a generative model, and define a new representation of the probabilistic bounding box through a spatial uncertainty distribution. Comprehensive experiments show that the proposed model reflects complex environmental noises in LiDAR perception and the label quality. Furthermore, we propose Jaccard IoU (JIoU) as a new evaluation metric that extends IoU by incorporating label uncertainty. We conduct an in-depth comparison among several LiDAR-based object detectors using the JIoU metric. Finally, we incorporate the proposed label uncertainty in a loss function to train a probabilistic object detector and to improve its detection accuracy. We verify our proposed methods on two public datasets (KITTI, Waymo), as well as on simulation data. Code is released at \url{https://bit.ly/2W534yo}.
\end{abstract}



\section{\textbf{Introduction}}\label{sec:introduction}
The availability of many real-world labelled datasets such as KITTI~\cite{geiger2012cvpr} and Waymo~\cite{sun2019scalability} is one of the key reasons behind the recent advancement in computer vision algorithms using deep learning in the field of autonomous driving~\cite{janai2017computer}. However, labels are not perfect. The process of annotating data can be error-prone due to human subjectivity or resource constraints. Ambiguity or uncertainty may also inherently exist in an object label due to sensor and environmental noises. Think about a 3D object detection task using LiDAR point clouds, where human annotators need to determine the object locations and extents, with LiDAR reflections only on the object's front surface. The leftmost figure of Fig~\ref{fig:introduction} depicts several ground truth bounding boxes (BBox) of cars in the LiDAR point clouds on the Bird's Eye View (BEV). Objects with dense LiDAR points are easy to be annotated, such as object 1 and 2, whereas objects with occlusion or being far-away from the ego-vehicle become difficult due to insufficient LiDAR observations, such as object 3 and 5. When training an object detection model, ignoring such uncertainty in ground truth labels may degrade the model's generalization capability, as each training data sample is encouraged to be fitted equally, even the one with remarkable noise. When evaluating the performance of a model, it is natural to downgrade the importance of noisy ground truths, because they do not necessarily reflect the true performance of the model. Therefore, modelling the uncertainty of labels is important for building robust and accurate object detectors in autonomous driving. Previous works have been focused on modelling uncertainty in class labels for image classification problem~\cite{sukhbaatar2014training,lawrence2001estimating,xiao2015learning,vahdat2017toward}. In this work, we focus on modelling uncertainty for object detection problem, which has not been widely explored to our knowledge. In particular, we are interested in the uncertainty in BBox labels, as their quality is critical in training and evaluating object detection. Largely-polluted BBox labels have been shown to degrade or even deteriorate the object detection accuracy~\cite{haase2019estimate}.

Uncertainty in ground truth labels should not only be considered in object detection models, but in evaluation metrics as well. Many evaluation metrics exist for object detection. For example, Intersection over Union (IoU) is the most common metric to measure localization accuracy in object detection. It is defined as the geometric overlap ratio between two (deterministic) BBoxes. Given the IoU values, several metrics to measure a detector's classification accuracy are proposed, such as Average Precision (AP)~\cite{everingham2010pascal} and Localization Recall Precision (LRP)~\cite{oksuz2018localization}. However, those metrics are designed only for deterministic object detection. They do not reflect the uncertainty estimation quality in probabilistic object detectors~\cite{feng2018towards,miller2017dropout,harakeh2019bayesod}, which provide predictive uncertainty estimates along with each detection. In this regard, the Uncertainty Error~\cite{miller2019evaluating} and the Probability-based Detection Quality (PDQ) metric~\cite{hall2018probabilistic} are proposed specifically for probabilistic object detection. All of the aforementioned metrics treat each ground truth label equally, without considering the uncertainty (or ambiguity) in the labeling process. As a consequence, existing evaluation metrics could not distinguish between noisy and clean ground truth label, and may not fully reflect the performance of deterministic and probabilistic object detectors.

\subsection*{\textbf{Contributions}}
In this work, we explicitly model the uncertainty of bounding box labels, which we call \textit{label uncertainty}, for object detection datasets with LiDAR point clouds. The label uncertainty is obtained by inferring through a generative model of LiDAR points. In this way, we can easily incorporate prior knowledge of sensor observation noises and annotation ambiguity into the label uncertainty. 

BBoxes are often labelled by their centroid positions, extents, and orientation. Therefore, the inferred label uncertainty is in a parameter space and is often high-dimensional. To visualize and represent the label uncertainty in the 3D space or on the LiDAR Bird's Eye View (BEV), we further propose a \textit{spatial distribution} of BBoxes. Using this spatial distribution, we verify that the proposed label uncertainty reflects many environmental noises in the LiDAR perception, which are related to distance, number of points, and typical L-shape observations. We also show that the label uncertainty reflects the quality of BBox labels in a dataset. Fig.~\ref{fig:introduction} visualizes the spatial distributions in several scenes.

Based on the spatial distribution of BBoxes, we propose a probabilistic extension of IoU, called \textit{Jaccard IoU (JIoU)}, as a new evaluation metric for object detection. The JIoU metric treats each BBox label differently according to its inferred uncertainty, and provides richer localization information than IoU. Using JIoU, we compare the detection performance among several state-of-the-art deterministic and probabilistic LiDAR-based object detectors.

Finally, we leverage the proposed label uncertainty to train a state-of-the-art probabilistic object detector. By incorporating the label uncertainty in the loss function, the detector's predictive uncertainty is regularized, resulting in a improved detection accuracy during testing.

We validate our methods in the KITTI~\cite{geiger2012cvpr} and Waymo~\cite{sun2019scalability} datasets, which are real-world datasets in autonomous driving, as well as in a simulator that can generate noisy-free ground truth labels.

In summary, the contributions of this paper are five-fold:
\begin{itemize}
    \item We build a generative model to infer the uncertainty inherent in BBox labels, and propose a spatial distribution to visualize label uncertainty.
    \item We propose a new evaluation metric for the object localization task, called Jaccard-IoU (JIoU). 
    \item We conduct extensive experiments to analyze the impact of parameters in label uncertainty, and justify the method regarding LiDAR observation noises and label quality. 
    \item We systematically compare several state-of-the-art deterministic and probabilistic LiDAR-based object detectors, using the JIoU metric.
    \item We leverage the proposed label uncertainty to improve a probabilistic object detector.
\end{itemize}

In the sequel, Sec.~\ref{sec:related_works} introduces related works, and Sec.~\ref{sec:methodology} presents the methodology of this paper. Afterwards, Sec.~\ref{sec:uncertainty_justification} conducts experiments to justify the proposed label uncertainty model, and Sec.~\ref{sec:experimental_result:jiou} evaluates deterministic and probabilistic object detectors with the JIoU metric. Finally, Sec.~\ref{sec:experimental_result:detection} introduces label uncertainty in training a probabilistic object detector, and Sec.~\ref{sec:conclusion} ends this paper with a brief conclusion and discussion.

\section{\textbf{Related Works}}
\label{sec:related_works}
 \subsection{\textbf{Modelling Label Uncertainty}} \label{subsec:related_works:label_noises}
Explicitly modelling ground truth label noises (or uncertainty) has been an active research topic in the field of computer vision and machine learning~\cite{frenay2013classification,algan2019image}.
While it is a common practice to assume independent Gaussian noises for the target label in regression~\cite{bishop2006pattern}, label noises in classification are much more complex. Class label noises can originate from insufficient data observations, lack of labellers' expertise or labelling biases, as well as data encoding and communication problems~\cite{frenay2013classification}. Incorrect labels, such as class flipping, may even deteriorate the model performance. Therefore, almost all methods in the literature focus on modelling class label noises and robust learning in image classification (e.g.~\cite{sukhbaatar2014training,lawrence2001estimating,xiao2015learning,vahdat2017toward}). In general, they identify noisy data using uncertainty models, and down-weight or correct those noisy labels to train classifier, using techniques such as label noise cleansing, dataset pruning, sample selection, and sample importance weighting. We refer interested readers to the survey~\cite{algan2019image} for detailed information.


In this work, we are interested in the regression uncertainty for ground truth bounding box (BBox) label parameters. They are largely affected by sensor observation noises (e.g. LiDAR point clouds), and noisy BBoxes have been shown to drastically degrade the object detection performance~\cite{haase2019estimate}. Only two works from Meyer \etal~\cite{meyer2019learning,meyer2019huber} are closely related to ours, which use simple heuristics to model BBox label uncertainty for object detection. In the first work~\cite{meyer2019learning}, the authors approximate the label uncertainty in LiDAR-based object detection by measuring the IoU value between a ground truth BBox and a convex hull of aggregated LiDAR points within that BBox. In the later work~\cite{meyer2019huber}, the authors consider label uncertainty in image-based object detection. They interpret the Huber loss, which is widely used in BBox regression, as the Kullback-Leibler Divergence between label uncertainty and predictive uncertainty. They place simple Laplacian noises on BBox parameters, and study their hyper-parameters based on intuitive understanding of label uncertainties. We diverge from these two works by explicitly infer the label uncertainty in LiDAR perception. We not only show how it reflects the typical L-shape observations in LiDAR point clouds and the label quality in a dataset, but also applies the label uncertainty in training and evaluating object detection networks. 

Our label uncertainty model can also be linked to the measurement models in LiDAR-based object tracking~\cite{granstrom2011tracking,scheel2018tracking,hirscher2016tracking}. We assume a measurement model which generates noisy LiDAR observations given a latent object state. Instead of inferring a probability distribution over the latent object state from multiple measurements over time and a prediction prior of the previous time-step, we infer a distribution over the latent object label with known mean from human annotators and measurements of a single time-step.

\subsection{\textbf{Standard Evaluation Metrics in Object Detection}} \label{subsec:related_works:evaluation_metrics}
In object detection, Intersection over Union (IoU) - the geometric overlap ratio between two BBoxes - has become the standard metric to measure localization accuracy. It is mainly used to threshold true positive predictions and false positives predictions, and has been extended as an auxiliary regression target to improve object detectors~\cite{rezatofighi2019generalized,zhou2019iou,zheng2020distance,jiang2018acquisition,he2020sassd}. 

Given a certain IoU threshold, Average Precision (AP) is derived as the standard metric to measure the object detection accuracy~\cite{everingham2010pascal}. AP is defined as the area under the precision-recall (PR) curve, approximated through numeric integration over a finite number of recall points. For example, in the KITTI object detection benchmark of ``Car'' class, The IoU threshold is set be $0.7$~\cite{geiger2012cvpr}, and AP is approximated by averaging the precision values at $11$ recall points. One drawback of the AP is that it does not directly reflect the localization performance in detection, since all predictions higher than the IoU threshold are treated equally. Instead, the MS COCO benchmark~\cite{lin2014microsoft} calculates AP averaged over several IoU thresholds ranging from $0.5$ to $0.9$; Oksuz \etal~\cite{oksuz2018localization} propose Localization Recall Precision (LRP) as a new evaluation metrics which incorporates the IoU score of each detection. The NuScenes object detection benchmark~\cite{caesar2019nuscenes} measures AP by matching objects based on their centroid euclidean distances rather than IoU scores, and proposes several metrics to specifically evaluate localization performance, including Average Translation Error (ATE), Average Scale Error (ASE), and Average Orientation Error (AOE). 

In this work, we replace IoU with JIoU to evaluate the performance of several state-of-the-art object detectors based on LiDAR point clouds. JIoU extends IoU by considering the label uncertainty in ground truth BBoxes, and provides richer localization information in LiDAR detection.

\subsection{\textbf{Probabilistic Object Detection Networks}} \label{subsec:related_works:uncertainty_estimation}
State-of-the-art object detectors are driven by deep learning approaches~\cite{janai2017computer}. They are often interpreted as point estimators with deterministic network weight parameters, as they classify objects with the softmax function, and regress BBox parameters using a single forward pass. In contrast to those ``deterministic'' object detectors, probabilistic object detectors (POD) aim to explicitly model predictive probability in network outputs, especially for the BBox regression task. One line of works exploits the sampling-based methods to do uncertainty estimation~\cite{miller2017dropout,feng2018towards,wirges2019capturing,miller2019evaluating}, such as MC-Dropout~\cite{Gal2016Uncertainty} and Deep Ensembles~\cite{lakshminarayanan2017simple}. They first collect prediction samples by performing multiple inferences with dropout or from an ensemble of object detectors, and measure the distribution statistics (e.g. Shannon entropy, mutual information). Another line of works~\cite{feng2018leveraging,feng2019can,he2019bounding,meyer2019lasernet,choi2019gaussian} directly assumes a certain probability distribution in network outputs, and uses output layers to regress the distribution parameters (e.g. mean and variance in a Gaussian distribution). POD have been applied to RGB cameras~\cite{le2018uncertainty,miller2017dropout,choi2019gaussian,he2019bounding}, LiDARs~\cite{feng2018towards,pan2020towards,meyer2019lasernet,wirges2019capturing}, and Radars~\cite{dong2020probabilistic}. They have been shown to improve the detection robustness in the open-set conditions~\cite{miller2017dropout}, reduce the labeling efforts in training~\cite{feng2019deep}, and enhance the detection accuracy~\cite{feng2018leveraging,he2019bounding,meyer2019lasernet,le2018uncertainty,choi2019gaussian}. In this work, we leverage the inferred label uncertainty to regularize a state-of-the-art LiDAR-based probabilistic object detector during training, and improve the detection performance during inference.

While the detection accuracy of POD is evaluated mainly by AP with IoU thresholds, their uncertainty estimation quality is studied by a variety of of metrics. For example, the uncertainty calibration plots~\cite{guo2017calibration,kuleshov2018accurate,feng2019can} evaluate how the predictive probabilities match the natural frequency of correct predictions. The Minimum Uncertainty Error~\cite{miller2019evaluating} reflects how the predictive probabilities can discriminate between true positive and false positive detections. The Probability-based Detection Quality (PDQ) proposed by Hall \etal~\cite{hall2018probabilistic} jointly measures the semantic uncertainty encoded by the classification probability vector and the spatial uncertainty encoded as the covariance matrices of BBox corners. The aforementioned uncertainty metrics only evaluate uncertainties from network outputs. Instead, our proposed JIoU takes account the label uncertainty, by comparing those output uncertainties of predictions with the spatial uncertainty distributions of ground truths.
\section{\textbf{Methodology}} \label{sec:methodology}
\subsection{\textbf{Problem Formulation}}
\label{sec:methodology:problem_formulation}
In standard object detection datasets for autonomous driving (such as KITTI~\cite{geiger2012cvpr} and Waymo~\cite{sun2019scalability}), ground truth labels often include object classes, denoted by a scalar $c$, and object poses and extents in the form of bounding boxes (BBoxes), whose parameters are denoted by a column vector of $d$ dimensions, $y\in \mathbb{R}^d$. Further denote $X_{\text{all}}=\{x_n\}_{n=1}^N$ as the set of all $N$ number of LiDAR points in a scan, and $x_n$ as one LiDAR point encoded by its position in the LiDAR coordinate frame ($x_n\in \mathbb{R}^3$ for 3D detection problem, and $x_n\in \mathbb{R}^2$ for 2D detection problem on the LiDAR Bird's Eye View (BEV) plane). \textit{Our target is to estimate the posterior distribution of the BBox parameters $y$, and extend the standard Intersection over Union (IoU) to a new metric called Jaccard-IoU (JIoU) that incorporates label uncertainty}. In this work, we demonstrate the method in the vehicle class, but it can be applied to other classes such as pedestrian and cyclist as well.

For 3D object detection problem, a BBox $B(y)\subset \mathbb{R}^d$ can be parameterized by its center location $c_1,c_2,c_3$, 3D extents (length $l$, width $w$, and heigth $h$), as well as orientation $r_y$, i.e. $y=[c_1,c_2,c_3,l,w,h,r]\in \mathbb{R}^7$. The set of all BBoxes is denoted as $S$. Then $Y: (S, \mathcal{A}, P_Y)\rightarrow \mathbb{R}^d$ is defined as the random variable of $y$ which maps the sample space $\mathcal{A}$ of BBoxes to the parameter space $\mathbb{R}^d$. It represents uncertain BBoxes with different probability $P_Y$. More specifically, a BBox label with uncertainty is denoted by $\overline{Y}$ and a BBox prediction with uncertainty is denoted by $\widehat{Y}$. 

\subsubsection*{\textbf{Assumptions}}
We usually have a prior knowledge of the object shape given its class $c$. Therefore, it is possible to infer the posterior distribution of $y$ given the object class $c$ and the LiDAR scan $X_{\text{all}}$, denoted by $p(y|X_{\text{all}}, c)$. For this purpose, we have the following three assumptions:
\begin{enumerate}
    \item The class label $c$ of an object is always correctly determined.
    \item The set of points belonging to an object are accurately segmented from the whole LiDAR scan by human annotators, with only a few outliers. 
    \item The Bbox label provided by human annotators, denoted by $\overline{y}$, is unbiased ($\overline{y}:=E(\overline{Y})=E\left(Y\right)$). Only the spread of $y$, or the variances and correlations in other words, need to be inferred. Note that the Bbox label is not updated by the posterior model, because this work intends to estimate uncertainty of human annotations, instead of correcting them.  
\end{enumerate}
Under these assumptions, the BBox label of an object $y$ is only conditioned on the set of LiDAR points belonging to this object, denoted by $x_{1:K} = \{x_k\}_{k=1}^{K}$, with $K$ the number of LiDAR points. Therefore, \textit{the posterior distribution} $p(y|X_{\text{all}}, c)$ \textit{mentioned above is reformulated as $p(y|x_{1:K})$ for notation simplicity}. We are especially interested in this posterior distribution, because human annotators may find it difficult to label BBoxes, when the LiDAR reflections are insufficient due to environmental and sensor noises such as occlusion and distance, as shown in Fig.~\ref{fig:introduction}.

\subsubsection*{\textbf{The Generation Process of LiDAR points}}
\begin{figure}[!tpb]
	\centering
	\subfigure[]{\includegraphics[width=0.47\linewidth]{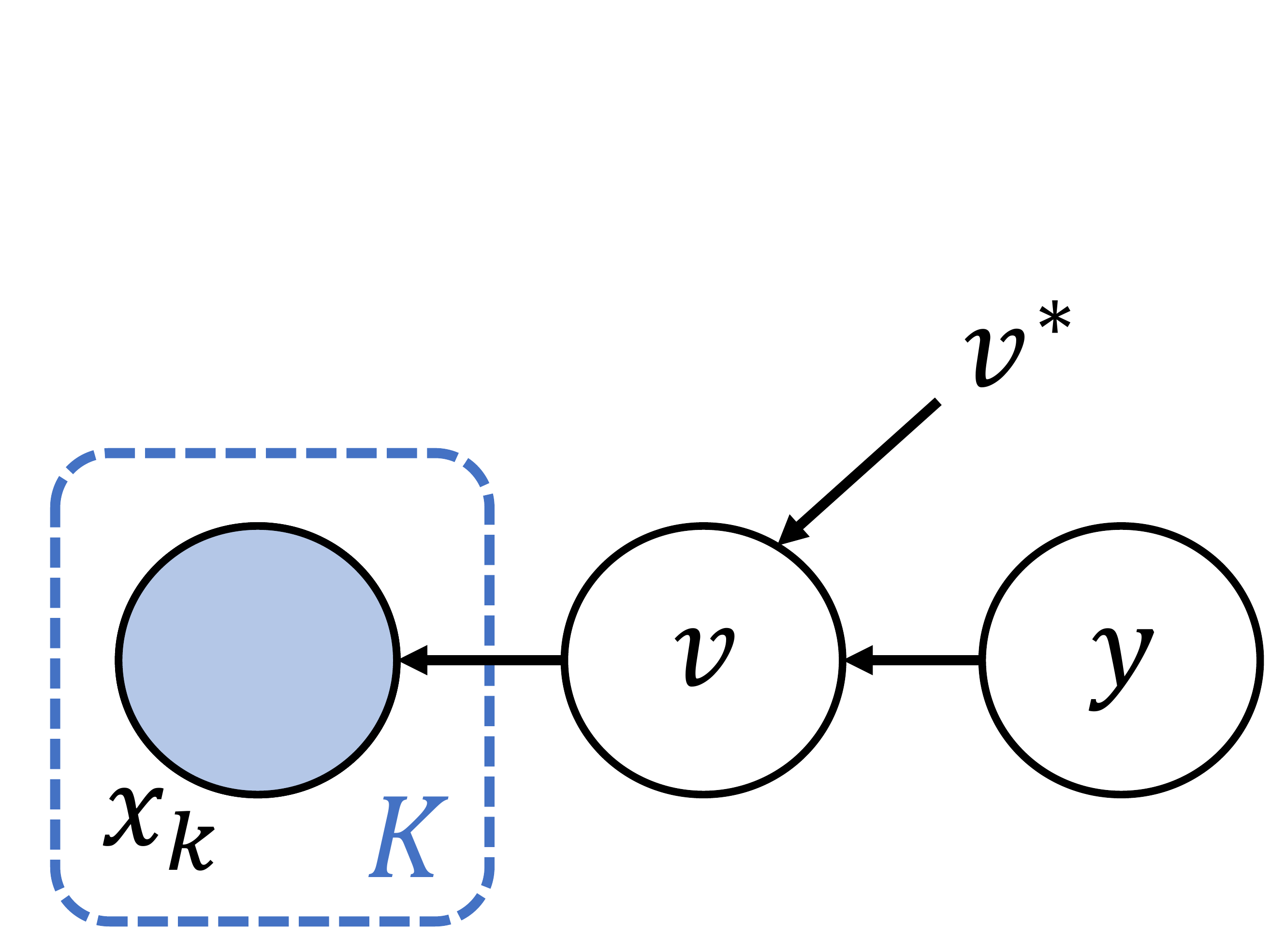}\label{fig:graphical_model_simple}}
	\hskip 12pt
	\subfigure[]{\includegraphics[width=0.47\linewidth]{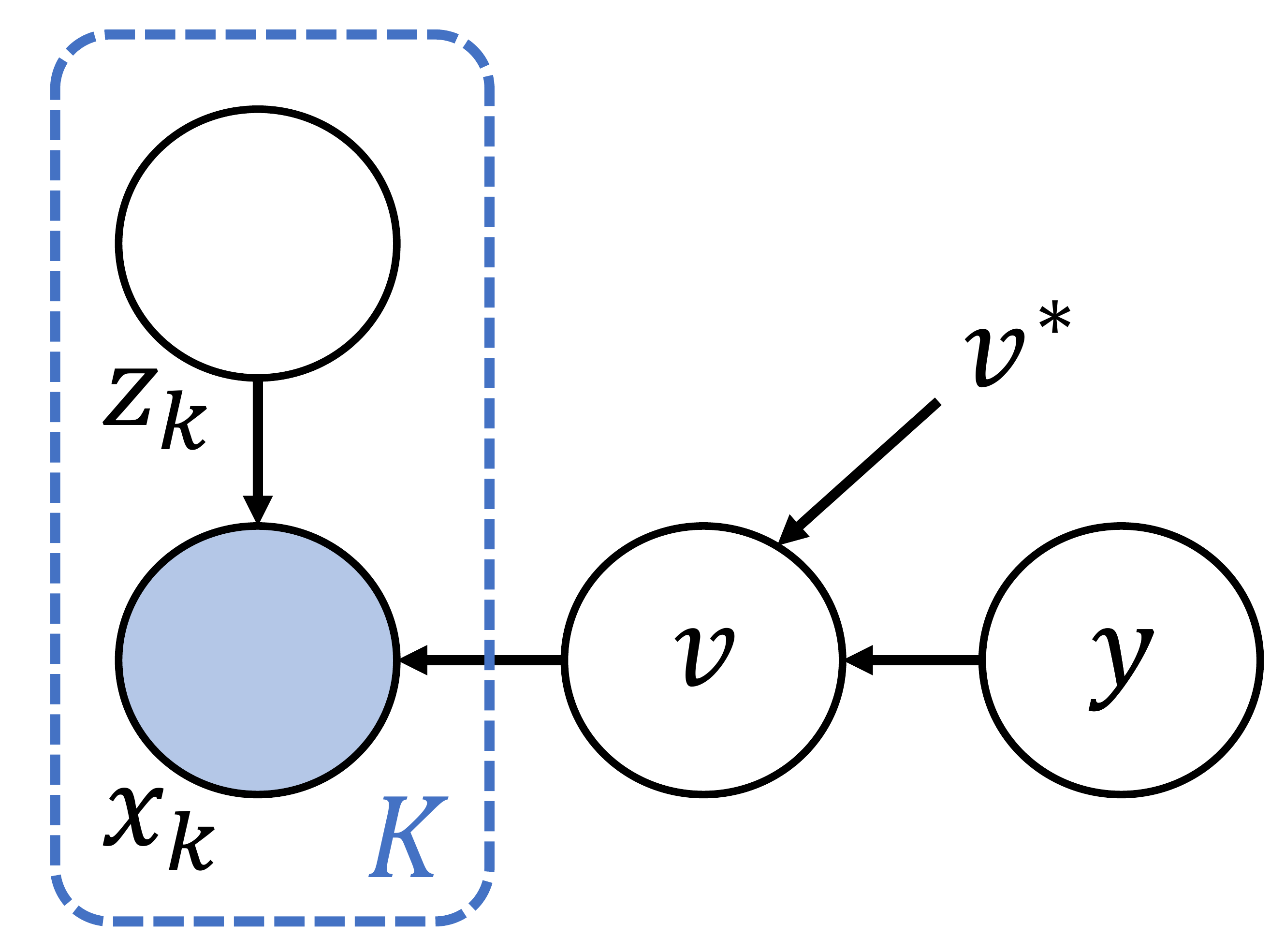}\label{fig:graphical_model_full}}
	\caption{Graphical representation of the proposed label uncertainty model. (a). The general model. (b). The model that introduces hidden categorical variable $z$.} \label{fig:graphical_model}
\end{figure}

We further assume that a LiDAR reflection $x$ of an object is generated by the object's surface with some noise. A location on the object's surface is denoted by $v\in B(y)$. In general, this \textit{unit} object's surface depends on the prior knowledge of the object's shape, which could simply be a BBox, an ellipsoid, or derived from some CAD models and point cloud rendering methods~\cite{fang2018simulating}. In this work, we focus on representing objects by their BBoxes parametrized by $y$. For example, recap that a 3D BBox is often encoded by its center location, 3D extents, and orientation, i.e. $y=[c_1,c_2,c_3,l,w,h,r_y]\in \mathbb{R}^7$. A location on this 3D BBox surface $v \in B(y) \subset \mathbb{R}^3$ is generated by a point on the \textit{unit} BBox $v^*\in B(y^*) = [-0.5,0.5]^3$ with an affine transformation.  The parameter $y^*$ of a \textit{unit} BBox is a fixed vector $[0,0,0,1,1,1,0]$. The affine transformation is consisted of scaling, rotation and translation given by:
\begin{equation}
    \label{eq:zy_render}
    v = R_y
\left[\begin{smallmatrix}
l& 0& 0\\
0& w& 0\\
0& 0& h
\end{smallmatrix}\right]
v^* + \left[\begin{smallmatrix}
c_1\\
c_2\\
c_3
\end{smallmatrix}\right]
\end{equation}
where $R_y$ is the rotation matrix of the orientation $r_y$. Similar to $Y$ and $y$, we define $V$ as the random variable of point $v$. Since the variable $V$ is dependent on $Y$ and $v^*$ following Eq.~\ref{eq:zy_render}, it can also be written as $V(v^*,Y)$ and $v(v^*,y)$, or $V(Y)$ and $v(y)$ for simplicity. The probabilistic graphical model LiDAR point generation process is shown in Fig.~\ref{fig:graphical_model_simple}.

\subsubsection*{\textbf{An Overview of the Proposed Methods}}
In the following sections, we will first introduce how to derive the posterior distribution $p(y|x_{1:K})$ via a variational Bayes method in Sec.~\ref{sec:methodology:label_uncertainty}. The label uncertainty is in the space of BBox parameters, which is often high-dimensional (for example, for 3D BBoxes, $y \in \mathbb{R}^7$ as introduced above). This makes it difficult to visualize and represent the uncertainty of a BBox. Therefore, in Sec.~\ref{sec:uncertainty} we further transform the posterior distribution $p(y|x_{1:K})$ into a spatial uncertainty distribution $p(u)$, with the vector $u$ being a location in the 3D space $u \in \mathbb{R}^3$ or in the LiDAR Bird's Eye View (BEV) space $u \in \mathbb{R}^2$. Such representation allows us to easily understand the spread of label uncertainty. Based on the spatial uncertainty distribution, we extend the standard IoU to a metric called Jaccard-IoU (JIoU) in Sec.~\ref{sec:methodology:JIoU}, which measures the object localization performance in a probabilistic manner. 

\subsection{\textbf{Label Uncertainty from Generative Model of Point Cloud}}
\label{sec:methodology:label_uncertainty}
To infer the posterior distribution $p(y|x_{1:K})$ of the random variable $\overline{Y}$ of BBox label given LiDAR observations $x_{1:K}$, we need to place an observation likelihood of LiDAR points $p(x_{1:K}|y, v)$, and a prior distribution of BBox label $p(y)$. The observation likelihood is conditioned on $v$, as we have assumed that a LiDAR point is generated from the BBox surface. The posterior is solved by the Bayes' rule:
\begin{equation} \label{eq:bayes1}
    p(y|x_{1:K}) \propto p(x_{1:K}|y, v)p(y).
\end{equation}
Concerning the observation likelihood, we assume that each LiDAR observation $x_{k}$ is independently generated with the same distribution from a object's surface, with:
\begin{equation} \label{eq:factorization}
p(x_{1:K}|y, v) = \prod_{k=1}^K p(x_{k}|y, v).
\end{equation}

In the following, we will first give a simple example of building an observation likelihood from the object's surface points, and solve the posterior distribution in the closed form. Afterwards, we introduce a more sophisticated observation likelihood based on a Gaussian Mixture Model (GMM), and solve its posterior distribution via Variational Bayes. For simplicity, this paper uniformly samples $v^*$ from the \textit{unit} BBox to generate the points on the actual object BBox surface $v$ following Eq.~\ref{eq:zy_render}.

\subsubsection{\textbf{A Simple Example}}
\begin{figure}[!tpb]
	\centering
	\includegraphics[width=6cm]{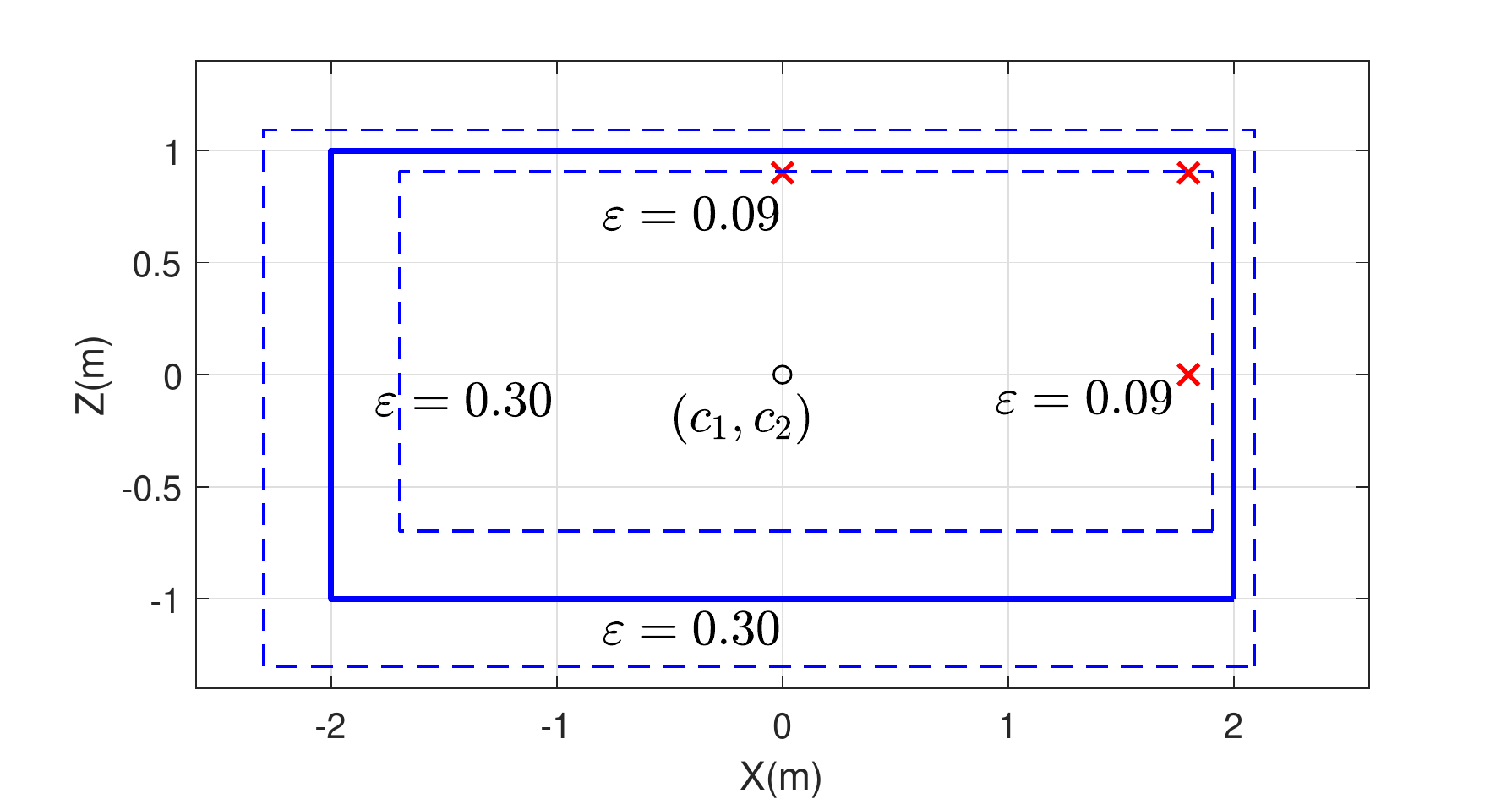}
	\caption{A simple example for the label uncertainty model.} \label{fig:demo1}
	\vspace{-3mm}
\end{figure}

Let us consider an axis-aligned 2D BBox on the Bird's Eye View (BEV) plane, with three LiDAR points $x_{1:3} = \{x_k\}_{k=1}^{3}$ shown by the red cross markers in Fig.~\ref{fig:demo1}, $x_1=[1.8,0]$, $x_2=[1.8,0.9]$, $x_3=[0,0.9]$. The BBox label parameters $y$ include the center positions $c_1,c_2$, length $l$, and width $w$, i.e. $y=[c_1,c_2,l,w] \in \mathbb{R}^4$. We assume that each LiDAR point $x_k \in \mathbb{R}^2$ is independently generated by its nearest point on the surface of the ground truth BBox denoted by $v_k$, with a Gaussian noise. In this example, the nearest surface points are $v_1=[c_1{+}0.5l,0]$, $v_2=[c_1{+}0.5l,c_2{+}0.5w]$, $v_3=[0,c_2{+}0.5w]$, and the observation likelihood is set to be:
\begin{equation} \label{eq:likelihood_simple}
p(x_k|y,v) = \mathcal{N}(x_k|v_k,\sigma^2 I),\quad \sigma=0.2
\end{equation}
where $\sigma$ is the standard deviation to model LiDAR noise, and $I$ an identity matrix. We further assume an (uninformative) Gaussian prior with large variance $p(y)=\mathcal{N}(y|\overline{y},100^2)$, with $\overline{y}$ being a unbiased mean value of the BBox label provided by human annotators. Combining Eq.~\ref{eq:likelihood_simple}, Eq.~\ref{eq:factorization} and Eq.~\ref{eq:bayes1}, the posterior distribution of this BBox label vector is Gaussian distributed, with:
\begin{equation} \label{eq:posterior_simple}
\begin{split}
    p(y|x_{1:3}) & \propto \prod_{k=1}^{3}{p(x_k|y, v_k)}p(y) \\
    & = \mathcal{N}\left(y\ |\ \cdot\ , \left[\begin{smallmatrix}
0.04& 0& {-}0.04& 0\\
0& 0.04& 0& {-}0.04\\
{-}0.04& 0& 0.06& 0\\
0& {-}0.04& 0& 0.06
\end{smallmatrix}\right]\right). \\
\end{split}
\end{equation}
Note that the mean value of $y$ is not updated, as we assume that it is given by human annotators accurately (cf. Sec.~\ref{sec:methodology:problem_formulation}). From Eq.~\ref{eq:posterior_simple}, we can observe that the derived covariance matrix has non-zero correlations. A singular value decomposition (SVD) of this covariance matrix shows that two edges of the BBox which have point cloud observations, namely $X=c_1+l/2$ and $Z=c_2+w/2$, have the smallest standard deviation $\epsilon=0.09$. The other two edges without observation, namely, $X=c_1-l/2$ and $Z=c_2-w/2$, have the largest standard deviation $\epsilon=0.30$. The confidence intervals of one standard deviation is shown by the blue dashed lines in Fig.~\ref{fig:demo1}. This result indicates that the BBox regions with more observations have less variances - a behaviour often observed in LiDAR perception.

\subsubsection{\textbf{Inferring Label Uncertainty with a Gaussian Mixture Model}} 
A more general way of modelling the observation likelihood $p(x_{k}|y, v)$ is some mixture model with categorical latent variable $z_k$. In this work, we employ a Gaussian Mixture Model (GMM) with $M$ components given by:
\begin{equation} \label{eq:GMM_generative_model}
    p(x_k|y, v) = \sum_{m=1}^{M}p\left(z_k=m\right)\mathcal{N}\left(v|v_{k,m}\ ,\ \sigma_m^2I\right) 
\end{equation}
with $z_k \in \{1,2,..M\}$. The component's mean values are assumed to be the $M$ nearest locations on the ground truth BBox surface to the LiDAR point $x_k$, denoted by $v_{k,m}$. The component's covariance $\sigma_m^2$ is an empirical value and related to the sensor noise and the confidence of the rendering method. This work assumes that they are all equal to a same value for simplicity, i.e. $\sigma_m^2=\sigma^2,\  \forall m\in \{1,...,M\}$, though a more complex model is possible depending on the rendering method. The graphical representation is shown in Fig.~\ref{fig:graphical_model_full}.

In this work, we solve the the posterior of GMM by variational Bayes (VB). A good practice of VB is the point registration method. The problem becomes to approximate the (actual) joint posterior distribution of $y$ and $z_{1:K}$, i.e. $p(y,z_{1:K}|x_{1:K})$, by a (simple) variational distribution $q(y,z_{1:K})$, through minimizing their Kullback-Leiber (KL) Divergence:
\begin{equation} \label{eq:VB_problem}
    D_{KL}(q||p) = \int_{z_{1:K}} q(y,z_{1:K})\log{\frac{q(y,z_{1:K})}{p(y,z_{1:K}|x)}}\ .
\end{equation}
Assuming $y,z$ are independent, the posterior distribution $p(y,z_{1:K}|x_{1:K})$ takes the form:
\begin{equation}\label{eq:joint_prob_factorization}
\begin{split}
   p(y,z_{1:K}|x_{1:K}) & = p(y|x_{1:K})p(z_{1:K}|x_{1:K}) \\
   & = p(y|x_{1:K})\prod_{k=1}^{K} p(z_k|x_k). \\
\end{split}
\end{equation}
Here, $p(y|x_{1:K})$ is the posterior distribution we would like to obtain, and $p(z_k=m|x_k)$ is the probability of registering the LiDAR point $x_k$ to a BBox surface point $v_{k,m}$.

Using the mean field method~\cite{blei2017variational}, the approximate distribution $q(y,z_{1:K})$ takes the form: $q(y,z_{1:K})=q(y)\prod_{k=1}^K q(z_k)$. In this way, the posterior distributions $p(y|x_{1:K})$ and $p(z_k|x_k)$ in Eq.~\ref{eq:joint_prob_factorization} are approximated by $q(y)$ and $q(z_k)$, respectively. The solution of the categorical distribution $q(z_k)$ is denoted by:
\begin{equation}
q(z_k=m)=\varphi_{k,m} \ \ \text{and}\ \ \sum_{m=1}^M \varphi_{k,m} =1.
\end{equation}
Using the mean field VB~\cite{blei2017variational}, the solution to $q(y)$ is given by:
\begin{equation} \label{eq:VB_solution}
\begin{aligned}
    q(y)&\propto \exp{\Big\{\log{p(y)}+\mathbb{E}_{z_{1:K}}\left[\log{p(x_{1:K}|z_{1:K},y)}\right]\Big\}}\\
    &\propto \exp\bigg\{\log{p(y)}-\frac{1}{2\sigma^2}\sum_{m=1}^{M}{\sum_{k=1}^{K}{\varphi_{k,m}||x_k-v_{k,m}(y)||^2}}\bigg\}.
\end{aligned}
\end{equation}
This solution results in a Gaussian distribution of $V(Y)$, but not a Gaussian distribution of $Y$, because $v$ and $y$ are not related by linear transformation. Instead, Eq.~\ref{eq:zy_render} can be written as below: 
\begin{equation}
\label{eq:zy_render_linear_form}
\begin{aligned}
v_{k,m}(y):&=J(v^*)\phi(y)\\
&=\begin{bmatrix}
1& 0& 0& v^*_1& 0& 0& -v^*_3& 0\\
0& 1& 0& 0& v^*_1& v^*_3& 0& 0\\
0& 0& 1& 0& 0& 0& 0& v^*_2\\
\end{bmatrix}\phi(y),\\
\phi(y):&=[c_1,c_2,c_3,l\cos{r_y},l\sin{r_y},w\cos{r_y},w\sin{r_y},h]^T,\\
v^* :&= [v_1^*, v_2^*, v_3^*]^T,
\end{aligned}
\end{equation}
where $\phi(y)$ is a feature vector from $y$, and the solved distribution $q(y)$ is a Gaussian distribution of $\phi(y)$. By assuming the prior distribution of to be a Gaussian of $\phi(y)$ with covariance $\Sigma_0$, the resulting covariance of $\phi(y)$ is
\begin{equation}
\label{eq:VB_solution_covariance}
\Sigma = \left(\Sigma_0^{-1}+\frac{1}{\sigma^2}\sum_{m=1}^{M}{\sum_{k=1}^{K}{\varphi_{k,m}J^T(v^*)J(v^*)}}\right)^{-1}.
\end{equation}
The point registration probability is given by:
\begin{equation} 
    \label{eq:registrator}
    \varphi_{k,m} = \frac{\exp\left(-\frac{1}{2\sigma^2}(x_k-\overline{v}_{k,m})(x_k-\overline{v}_{k,m})^T\right)}{\sum_{m=1}^M{\exp\left(-\frac{1}{2\sigma^2}(x_k-\overline{v}_{k,m})(x_k-\overline{v}_{k,m})^T\right)}},
\end{equation}
with $\overline{v}_{k,m}$ being a surface point on the ground truth BBox, whose label values $\overline{y}$ are provided by human annotators. Note that it is different from $v_{k,m}$, which is dependent on the $y$ and is iteratively updated using the EM algorithm in Eq.~\ref{eq:VB_solution}.

Finally, the solution to the observation noise $\sigma$ is given in the form:
\begin{equation}\label{eq:parameter_EM}
    \sigma^2=\frac{1}{MD}\sum_{m=1}^{M}{\sum_{k=1}^{K}{\varphi_{k,m}(x_k-\overline{v}_{k,m})(x_k-\overline{v}_{k,m})^T}},
\end{equation}
with $D$ being the dimension of BBox parameters. In practice, this observation noise can also be chosen empirically based on the prior knowledge of LiDAR points.

The results of the proposed variational Bayes method from Eq.~\ref{eq:VB_solution}, Eq.~\ref{eq:registrator}, and Eq.~\ref{eq:parameter_EM} are used to calculate the label uncertainty of vehicle objects based on LiDAR point clouds. 

\subsection{\textbf{Spatial Uncertainty Distribution of Bounding Boxes}}
\label{sec:uncertainty}
The posterior distribution (or label uncertainty) presented in the previous section is in the space of BBox parameters, which is often high-dimensional depending on the BBox encoding methods. In this section, we propose to transform this label uncertainty distribution into a spatial uncertainty distribution of BBoxes denoted by $p(u)$, with $u$ being a location in 3D ($u \in \mathbb{R}^3$) or BEV ($u \in \mathbb{R}^2$). It provides a visualization of label uncertainty, which can also be used in the JIoU defined in Sec.~\ref{sec:methodology:JIoU}.

The idea of probabilistic box representation is proposed by the Probability-based Detection Quality (PDQ) in~\cite{hall2018probabilistic} for 2D axis-aligned BBoxes in images. The resulting spatial distribution $P(u)\in \left[0,1\right]$ denotes the probability of $u\in \mathbb{R}^2$ within a Bbox, which is equal to the probability of $Y$ such that the BBox contains $u$, i.e. $P_Y(\{y|u\in B(y)\})$. Then the probability is calculated by: 
\begin{equation}
    \label{eq:Probabilistic_object}
    P_{PDQ}(u) := \int_{\{y|u\in B(y)\}}{p_Y(y|x)dy},
\end{equation}
where $Y$ is the random variable of the BBox parameters and $B(y)$ is the BBox given parameters $y$. More formally, the spatial distribution defined by PDQ is a random field on the Euclidean space indexed by $u\in \mathbb{R}^n$, $n=2$ or $3$, and each random variable is binary-valued and defined as whether the point $u\in B(y)\subset \mathbb{R}^n$.  Eq.~\ref{eq:Probabilistic_object} is easy to calculate for axis-aligned BBoxes but hard for rotated boxes, because it has to integrate over the space of $y$, which is 7 dimensional for 3D. This motivates us to propose a new way of efficiently calculating spatial distribution denoted by $p_G$. A transformation in the integral gives another expression as a probabilistic density function (PDF): 
\begin{equation}\label{eq:Generative_object}
    \begin{aligned}
    p_G(u)&:=\int_{v^*\in B(y^*)}p_{V\left(v^*,Y\right)}\left(u\right)dv^*\\
    &=\int_{\{y|u\in B(y)\}}{\frac{1}{A(y)} p_Y(y|x)dy},
    \end{aligned}
\end{equation}
where $V(v^*,Y)$ is defined in Eq.~\ref{eq:zy_render}, and $V$ and $Y$ are random variables. $v^*$ is added as an argument because we need to integrate over $v^*$. Details of proof is provided in Appendix~\ref{appendix:proof_spatial_uncertainty}. The label uncertainty derived from Sec.~\ref{sec:methodology:label_uncertainty} gives a Gaussian distribution of $V(v^*,Y)$, which leads to an easy calculation of $p_G(u)$. The calculation of spatial distribution, when $V(v^*,Y)$ is not explicitly given such as the predictions with uncertainty from~\cite{feng2019can}, is included in Appendix~\ref{appendix:howto_spatial_uncertainty}.

The proposed definition of spatial distribution $p_G$ is slightly different from $P_{PDQ}$ by scaling the density with the size $A(y)$ of the BBox $B(y)$. The scaling factor enables the transformation to a integral over distributions generated by points $v^*$ inside the unit box $B(y^*)$, i.e. $A\left(y^*\right){=}1$. Therefore, it has a significant advantage of reducing the dimension of the integral from 7 to 3 for 3D. Besides, $\int_u p_G(u){=}1$ if it is integrated on spatial points $u$, while $P_{PDQ}(u)$ is not normalized. The shapes of their distribution differ only when the object size is uncertain and the proposed $p_G(u)$ tend to be more concentrated.

\subsection{\textbf{JIoU: A Probabilistic Generalization of IoU}}
\label{sec:methodology:JIoU}
Evaluating the quality of detection results requires intuitive scalar values. The most commonly used metric is Intersection over Union (IoU), which measures the geometric overlap ratios between two BBoxes. However, IoU only measures deterministic BBoxes, for example, parameterized by $y_1$ and $y_2$. In contrast, probabilistic predictions or labels provide the distributions of random variables $Y_1$ and $Y_2$ of two Bboxes, and an evaluation metric that is defined on distribution is desired. In this section, Jaccard IoU (JIoU) is proposed as a natural extension of IoU on probabilistic Bboxes. Inspired by the definition of the probabilistic Jaccard index~\cite{moulton2018maximally}, the JIoU metric is given by:
\begin{equation}
    \label{eq:JIoU_def}
    \text{JIoU}(Y_1, Y_2) := \int_{R_{1}\cap R_{2}} {\frac{du}{\int_{R_1 \cup R_{2}}{\max\left(\frac{p_1(u')}{p_1(u)},\frac{p_2(u')}{p_2(u)}\right)du'}}}\ ,
\end{equation}
where $p_1,p_2$ are the spatial uncertainty distributions of random variables $Y_1$ and $Y_2$ given by either Eq.~\ref{eq:Probabilistic_object} or Eq.~\ref{eq:Generative_object}. Also, $u, u'$ are arbitrary locations in the 3D or BEV space, and $R_1,R_2$ are the supports of $p_1,p_2$, respectively. JIoU ranges within $[0,1]$, and is maximized only when two Bboxes have the same locations, same extents, as well as same spatial uncertainty distributions. The computational complexity is $O(NlogN)$ using Eq.~3 of~\cite{moulton2018maximally}, if $N$ points are sampled from $R_1\cup R_2$. Specifically, JIoU simplifies to IoU when two Bboxes are deterministic (i.e. the distributions of $Y_1, Y_2$ are delta functions, $R_1, R_2$ become BBoxes and $p_1, p_2$ become uniform distributions inside their boxes). 

\begin{figure}[h!]
	\centering
	\subfigure[$\text{JIoU}{=}0.1$ using $P_{PDQ}$]{\includegraphics[width=3.5cm]{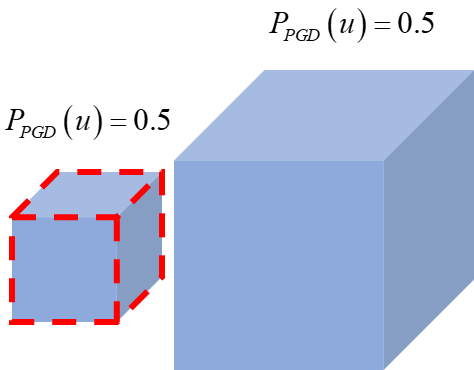}}
	\hspace{3mm}
	\subfigure[$\text{JIoU}{=}0.5$ using $p_{G}$]{\includegraphics[width=3.5cm]{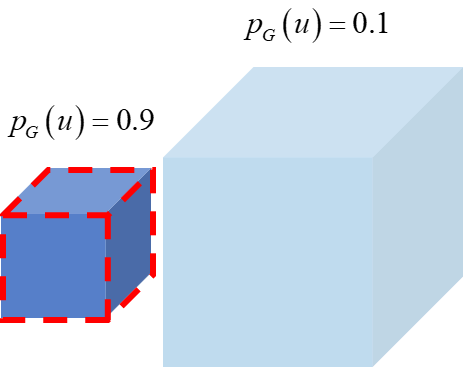}}
	\caption{Spatial distributions of discrete $\overline{Y}$ with two possible values in blue and predicted box $\hat{Y}=\hat{y}$ in dashed red line.}
	\label{fig:JIoU_with_different_distribution}
\end{figure}

The representation of spatial distribution $p_G(u)$ proposed in Sec.~\ref{sec:uncertainty} is more reasonable than $P_{PDQ}(u)$, when JIoU is used as an evaluation metric (although JIoU accepts either one). An example is given in Fig.~\ref{fig:JIoU_with_different_distribution} with prediction $\widehat{Y}$ and label $\overline{Y}$. Here $\overline{Y}$ is a discrete random variable with two values of equal probability, i.e. two possible BBoxes. The prediction $\hat{Y}$, which is deterministic, only fits one of the box. It is desired that $\text{JIoU}=0.5$ because the prediction has matched one of the two possible labeled BBoxes. If $P_{PDQ}$ is used, the spatial distribution of the label is a constant $0.5$ and $\text{JIoU}(\overline{Y},\hat{Y}){\approx} 0$ because prediction only covers the small box. If $p_{G}$ is used, $\text{JIoU}(\overline{Y},\hat{Y}){=}0.5$, no matter what the size difference is. 
\section{\textbf{Justification of the Uncertainty Model}} \label{sec:uncertainty_justification}
In this section, we conduct experiments to justify our proposed method which builds a Gaussian Mixture Model (GMM) to infer posterior distribution of bounding box (BBox) labels. First, we study the impact of parameters on label uncertainty (Sec.~\ref{sec:uncertainty_justification:choice_of_parameters}). Then, we study the complex environmental noises reflected by the uncertainty model, such as typical ``L-shape'' in LiDAR perception, occlusion, distance, and LiDAR point density (Sec.~\ref{sec:uncertainty_justification:lidar_observation}). Finally, we explore the behaviours of label uncertainty at different label quality levels with the help of a simulator (Sec.~\ref{sec:uncertainty_justification:label_quality}). 

\subsection{\textbf{Setup}}\label{sec:uncertainty_justification:setup}
We quantify label uncertainty \textit{at an arbitrary location in the space} by its variance derived from the uncertainty model, or by its spatial uncertainty distribution value (i.e. the probability that the location belongs to the object). We also quantify label uncertainty \textit{for a ground truth BBox} by its JIoU score (called ``JIoU-GT''). A JIoU-GT score is measured between a deterministic ground truth BBox (which corresponds to a uniform distribution in the JIoU calculation) and its spatial uncertainty distribution derived from the label uncertainty model (which reflects the uncertainty inherent in the ground truth label). A larger JIoU-GT value indicates smaller label uncertainty. In addition, we visualize the label uncertainty by its spatial distribution on the LiDAR Bird's Eye View (BEV). We choose $M=3$ in the GMM model (Eq.~\ref{eq:GMM_generative_model}), which is found enough to capture detailed distribution of LiDAR points. Furthermore, we conduct experiments for BBoxes on the LiDAR Bird's Eye View (BEV) plane, and ignores their vertical positions and heights. We are especially interested the BEV representation, because most objects of interest in autonomous driving are constrained on the ground plane, and there are only a few variations in object's heights, as discussed in~\cite{yang2018pixor}. Nevertheless, our proposed can be directly applied to 3D BBoxes as well.

In Sec.~\ref{sec:uncertainty_justification:choice_of_parameters} and Sec.~\ref{sec:uncertainty_justification:lidar_observation}, we evaluate the uncertainty model on the KITTI object detection benchmark~\cite{geiger2012cvpr} and the Waymo open dataset (WOD)~\cite{sun2019scalability}. Both datasets provide LiDAR point clouds and 3D object labels from human annotators. We use the ground truths of ``Car'' and ``Van'' categories in the KITTI training dataset, with $7481$ data frames and approximately $30$K objects. In WOD, we select the training data drives recorded in San Francisco, and down-sample the frames by a factor of $10$ in order to remove redundancy between sequential frames. WOD only provides the ``Vehicle'' class, without distinguishing among ``Motorcycle'', ``Car'', ``Van'', or ``Truck'' classes. In order to directly compare WOD with KITTI, we further extract small vehicle objects in WOD by thresholding objects with length within $\unit[3]{m}-\unit[6.5]{m}$, similar to ``Car'' and ``Van'' classes in KITTI. As a result, we collect $7545$ frames and over $150$K ground truths using the Waymo open dataset. 

\subsection{\textbf{Choices of Parameters for the Label Uncertainty}}\label{sec:uncertainty_justification:choice_of_parameters}
As introduced in Sec.~\ref{sec:methodology:label_uncertainty}, the inference of label uncertainty in the posterior distribution $p(y|x_{1:K})$ incorporates the prior knowledge of LiDAR observation noise in the observation likelihood model $p(x_k|y,v)$, as well as human annotation noise in the priors $p(y)$. In this section, we show the impact of LiDAR observation noise and human annotation noise on the proposed uncertainty model. 
\begin{figure}[!tpb]
	\centering
	\begin{minipage}{1\linewidth}
	\centering
		\includegraphics[width=1\linewidth]{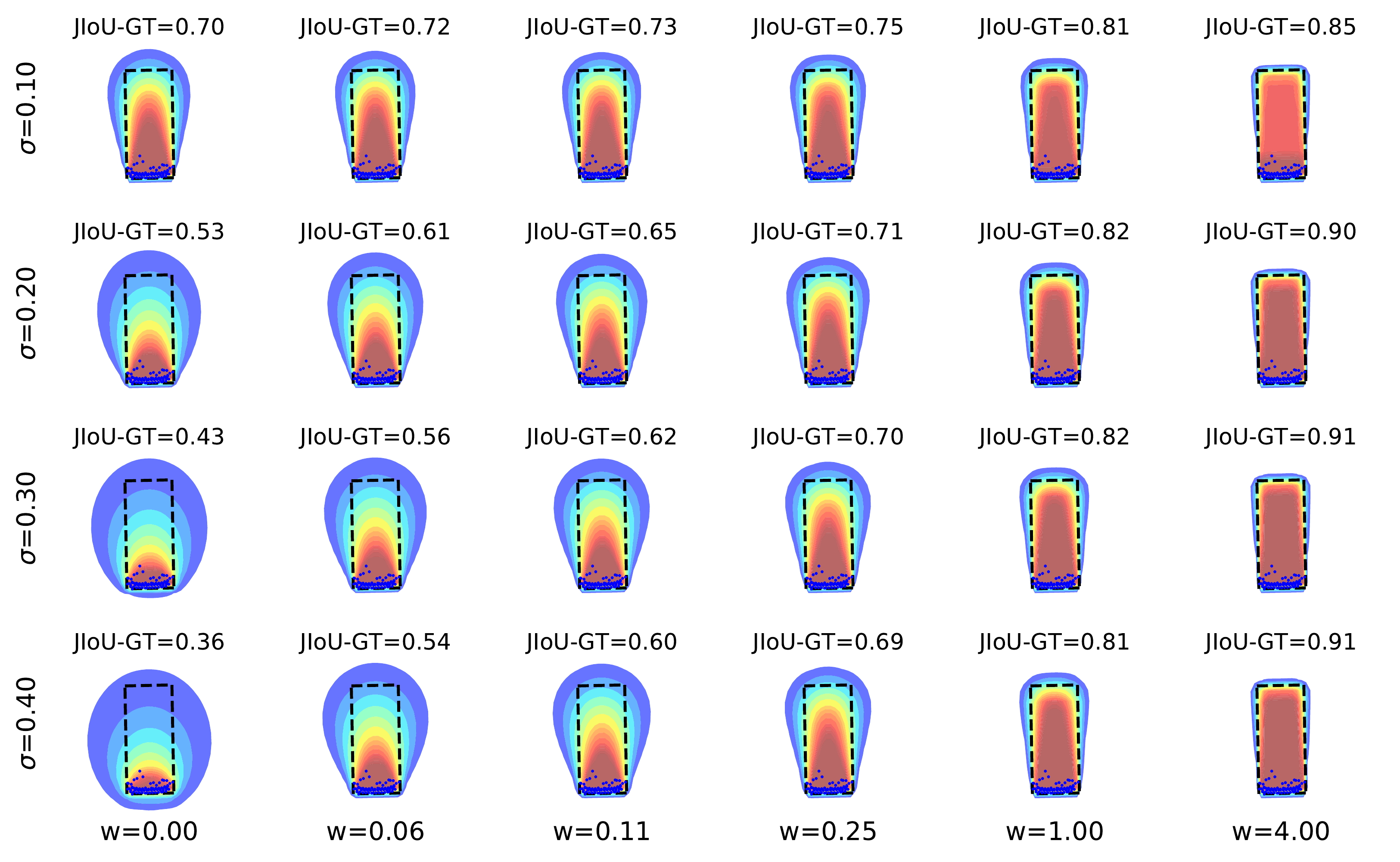}
		\caption{Influence of LiDAR measurement noise and prior distribution on the spatial uncertainty distribution. Better view with magnificence.} \label{fig:exp_parameter_influence}
	\end{minipage}
\end{figure}
\begin{figure*}[!pbt]
	\centering
	\begin{minipage}{0.92\linewidth}
		\centering
		\includegraphics[width=1\linewidth]{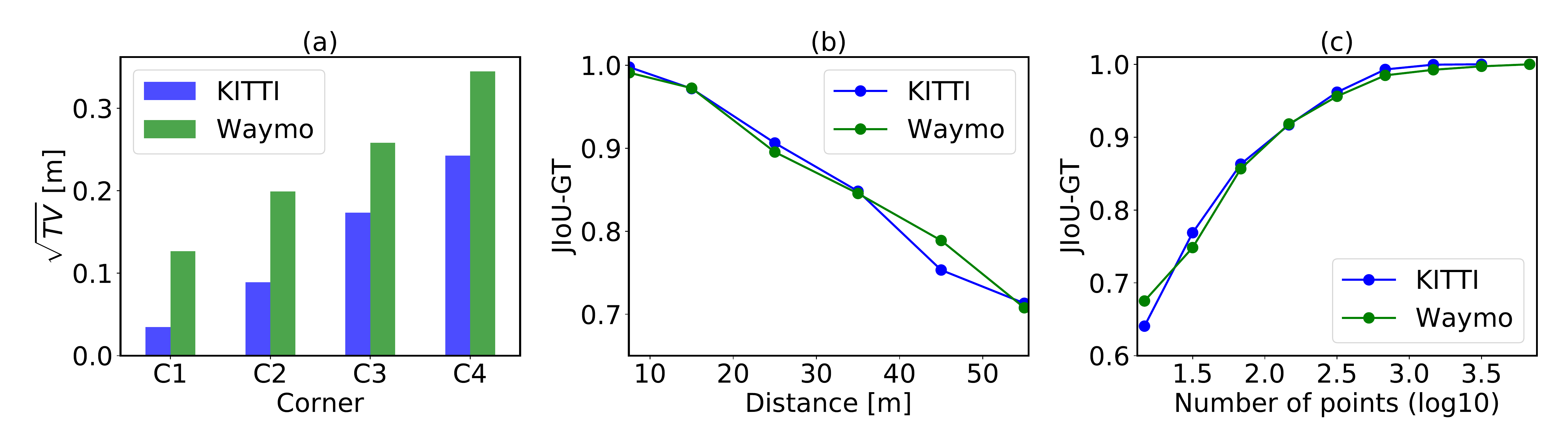}
		\caption{A study of the behaviours of the proposed uncertainty model using KITTI~\cite{geiger2012cvpr} and Waymo~\cite{sun2019scalability} datasets. (a). The average total variances~\cite{feng2018leveraging} of ground truth BBox corners. The variances are inferred by the proposed uncertainty model. Corners (C1-C4) are sorted by their distances to the ego-vehicle from the closest to the farthest. (b). The average JIoU-GT scores w.r.t. the object distance. (c). The average JIoU-GT scores w.r.t. the logarithmic number of LiDAR observations within BBoxes.}\label{fig:spatial_uncertainty}
	\end{minipage}
\end{figure*}

\begin{figure*}[!pbt]
	\centering
	\begin{minipage}{0.92\linewidth}
		\centering
		\includegraphics[width=1\linewidth]{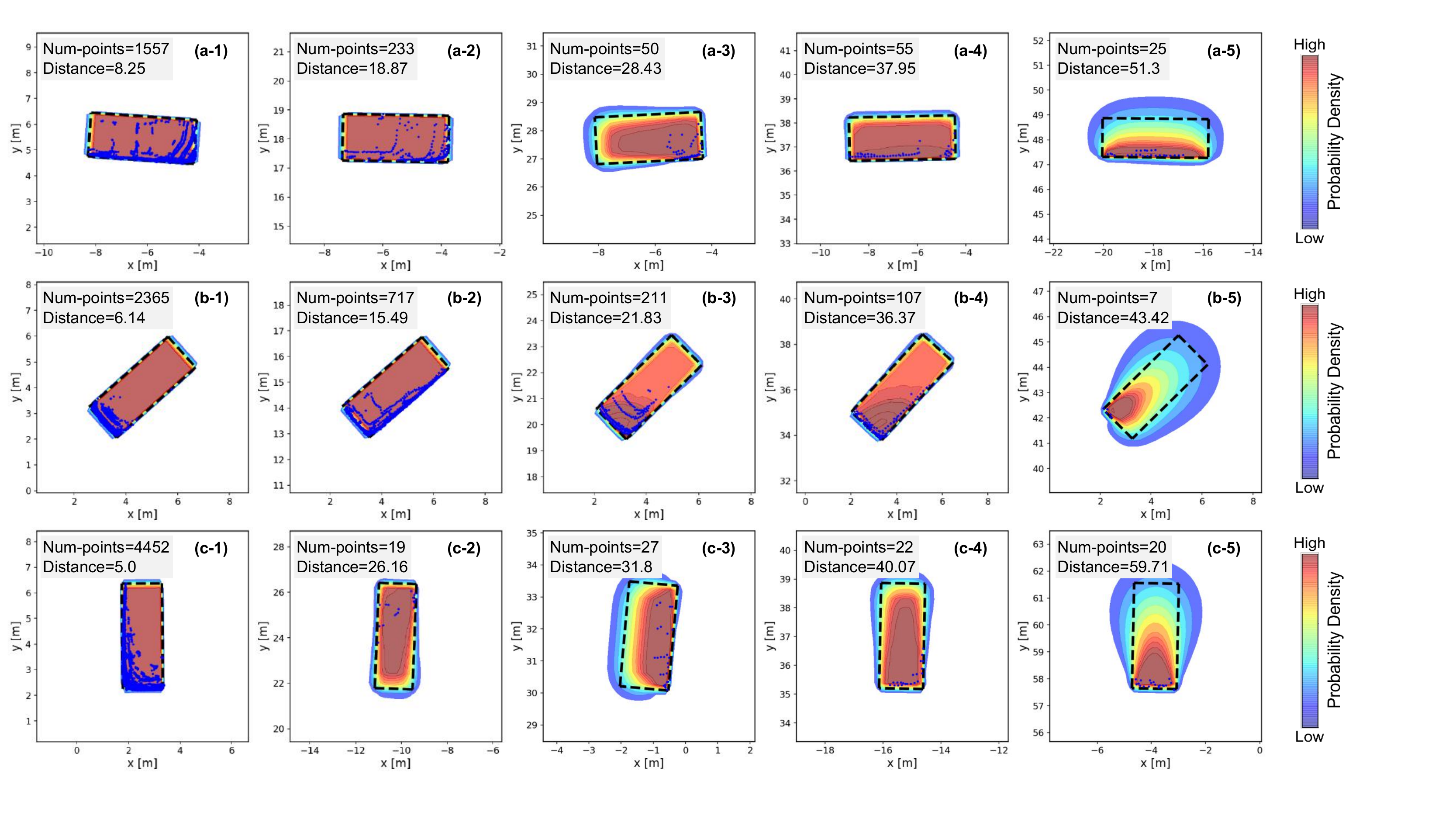}
		\caption{Several ground truth BBoxes on BEV and their inferred spatial uncertainty distributions. The horizontal axis x corresponds to the lateral distance, the vertical axis y the longitudinal distance. Better view with magnificence.}\label{fig:spatial_uncertainty_distribution}
	\end{minipage}
\end{figure*}
We examine the standard deviation of LiDAR observation noise denoted by $\sigma$, and the weighted covariance matrix of $p(y)$ denoted by $\Sigma_0$. To define $\Sigma$ empirically, we first parameterize the BEV BBox by its centroid positions on the BEV plane $c_1, c_2$, length $l$, width $w$, and the yaw angle $r$, i.e. $y=[c_1,c_2,l,w,r]$. Then, we calculate the variances of all object ground truth BBoxes in a dataset, which are considered as the prior knowledge of human annotation noise. Finally, we weight all variances by a factor $w$ to control the strength of prior knowledge. For example, for objects of the ``Car'' class in the KITTI dataset, the covariance matrix is given by: 
\begin{equation}
\Sigma_0 = \frac{1}{w}\left[\begin{smallmatrix}
    0.44^2& 0& 0& 0 & 0\\
    0& 0.11^2& 0& 0 & 0\\
    0& 0& 0.25^2& 0 & 0\\
    0& 0& 0& 0.25^2 & 0\\
    0& 0& 0& 0 & 0.17^2
\end{smallmatrix}\right].
\end{equation}
Fig.~\ref{fig:exp_parameter_influence} shows the change of spatial uncertainty distribution and JIoU-GT scores for a BBox label with respect to $\sigma$ and $w$. In this example, there exist LiDAR observations only on the front surface of the object, leading to a spatial uncertainty distribution with high density in the front surface and low density on the backside. When increasing $w$ (from left to right plots) or decreasing $\sigma$ (from bottom to upper plots), we observe that the spatial uncertainty distribution is ``sharpened'' with higher JIoU-GT scores. This is because a larger $w$ value indicates the smaller variances in the prior distribution $p(y)$, and thus more confidence in the BBox values provided by human annotators. A smaller $\sigma$ value indicates smaller noise in the LiDAR perception, which reduces uncertainty. Note that with $w=0$, the covariance matrix $\Sigma_0$ goes to infinitive, resulting in a non-informative prior. More objects are visualized in the appendix (Fig.~\ref{fig:exp_parameter_influence_appendix}). Interestingly, though the same behaviours of uncertainty distribution from $\sigma$ and $w$ can be seen for all objects in a dataset, we observe that the label uncertainty for objects with dense LiDAR observations are less affected by the choice of parameters. They have  quasi-uniform distributions, regardless of the changing $\sigma$ and $w$ values (eg. Fig.~\ref{fig:pscale_000186_0} in the appendix).

Instead of empirically choosing the value of $\sigma$, it can be estimated by the EM algorithm~\cite{lawrence2001estimating} following Eq.~\ref{eq:parameter_EM}. The resulting $\sigma$ is $0.2$m for KITTI and $0.3$m for Waymo. Eq.~\ref{eq:parameter_EM} implies that these results are very close to the root mean square of distances between LiDAR points and BBox surface. The estimated value of $\sigma^2$ includes both the measurement noise of the LiDAR sensor and the approximation error of a BBox to the actual surface of a car.

\subsection{\textbf{Label Uncertainty and LiDAR Observations}}\label{sec:uncertainty_justification:lidar_observation}
In this section we study the behaviours the proposed uncertainty model with complex environmental settings in the LiDAR perception.

First, Fig.~\ref{fig:spatial_uncertainty}(a) shows the standard deviation of the total variances (TV) for four corners in the BEV BBox, averaged over all objects in the KITTI and Waymo datasets. The TV scores are calculated by the proposed spatial distribution, and the corners (C1-C4) are sorted by their distance to the ego-vehicle in the ascending order. The TV scores are increasing from the nearest corner C1 to the farthest corner C4. This observation corresponds to the intuition of L-shapes~\cite{zhang2017efficient} widely used for LiDAR-based vehicle detection and tracking: The nearest corner of the L-shape, which usually receives dense LiDAR observations, is regarded more reliable than the distant and occluded corners by our proposed uncertainty model. Note that the TV scores in the Waymo dataset are higher than those in the KITTI dataset, because the Waymo dataset has much more objects with sparse LiDAR points and long distance compared to KITTI (cf. Fig.~\ref{fig:kitti_waymo_data_distribution} in the appendix). 

Then, Fig.~\ref{fig:spatial_uncertainty}(b) shows the evolution of label uncertainty in terms of JIoU-GT w.r.t the object distance. We observe a decrease of JIoU-GT scores with larger distance in both datasets, indicating that distant objects are regarded difficult to be labelled in our uncertainty model. Fig.~\ref{fig:spatial_uncertainty}(c) shows the change of JIoU-GT scores w.r.t. the number of LiDAR points within BBoxes (at the $\log_{10}$ scale). The JIoU-GT scores increases (i.e. smaller uncertainty) given more dense LiDAR points.

Finally, Fig.~\ref{fig:spatial_uncertainty_distribution} shows qualitatively $15$ ground truth objects on the BEV and their inferred spatial uncertainty distributions. We divide them into three groups according to the rotation (row a: $90$ or $270$ degrees, row b: approx. $45$ degrees, row c: $0$ or $180$ degrees). Objects in each group are ordered based on their increasing distance from left to right. The spatial uncertainty distributions become ``softened'' with increasing distance and reducing number of LiDAR points (from the left to the right plots). They also show the ``L-shape'' behaviours such as in Fig.~\ref{fig:spatial_uncertainty_distribution}(b-3), (b-4) and (c-4). In Fig.~\ref{fig:spatial_uncertainty_distribution}(a-3) and (b-5), objects are highly occluded, and only small parts have LiDAR reflections. Correspondingly, the inferred uncertainty is low in regions with LiDAR points, and high outside.

In conclusion, our proposed method captures reasonable uncertainty inherent in ground truth BBoxes. The uncertainty shows the ``L-shape'' behaviours, and is affected by distance, occlusion, and LiDAR density.

\subsection{\textbf{Label Quality Analysis}}\label{sec:uncertainty_justification:label_quality}
\begin{figure}[t]
	\centering
	\begin{minipage}{0.92\linewidth}
		\centering
		\includegraphics[width=1\linewidth]{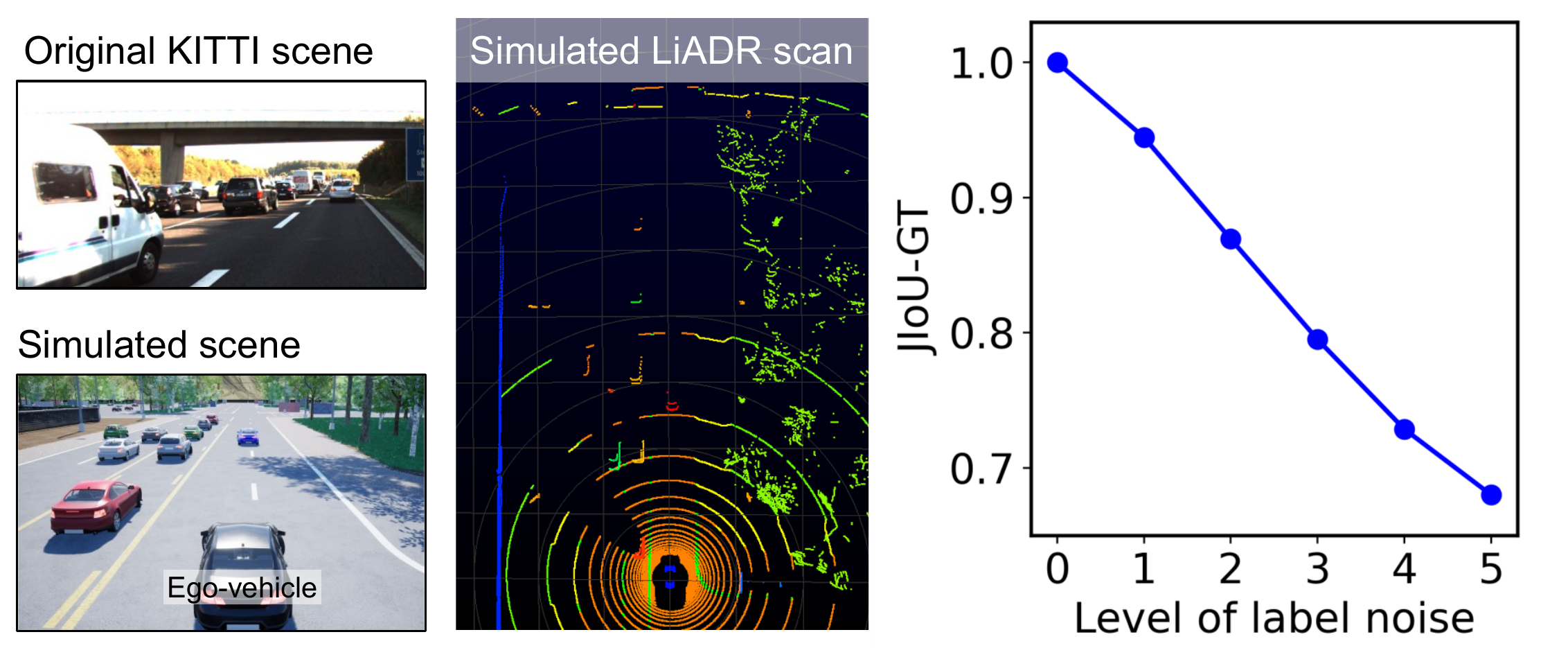}
		\caption{Left: an original highway scene from the KITTI dataset~\cite{geiger2012cvpr} and its corresponding simulation using the monoDrive simulator~\cite{monodrive}. Right: Illustration of the averaged JIoU-GT scores w.r.t increasing level of label noise. Level-0 refers to label without any noise.}\label{fig:simulation_data}
	\end{minipage}
\end{figure}

Finally, we study how the proposed label uncertainty model reflects the labelling quality of BBoxes. Ideally, the method should depict higher uncertainty values for noisy labels than accurate labels. However, a direct to study in public object detection datasets is difficult, because they do not provide the ``ground truth'' of label noises. Instead, we leverage a simulator to build typical driving scenarios with noisy-free object's ground truths and then generate synthetic label noises. To this end, we first collect nearly $300$ scenes for the KITTI~\cite{geiger2012cvpr} and Waymo~\cite{sun2019scalability} datasets, and build their corresponding simulated scenes with LiDAR scans by the monoDrive simulator~\cite{monodrive}, as shown in Fig.~\ref{fig:simulation_data}(left). Afterwards, we synthetically add Gaussian noises on each object's label (centroid positions, length, width) with increasing level of standard deviations up to $\unit[1]{meter}$. Finally, we use the JIoU-GT metric to represent label uncertainty. The JIoU-GT score for a BBox label with added noise is normalized by its corresponding score without noise (Level-0). Fig.~\ref{fig:simulation_data}(right) depicts the mean values of JIoU-GT scores averaged over all simulated objects w.r.t. increasing label noises. We observe a consistent decrease of JIoU-GT scores (i.e. higher label uncertainty) with higher noise label (i.e. worse label quality), showing that the proposed model is able to reflect label quality.

\section{\textbf{Using JIoU to Evaluate Object Detectors}}\label{sec:experimental_result:jiou}
In object detection, IoU is the standard metric to measure localization accuracy and to distinguish between true positive and false positive predictions. For instance, the KITTI object detection benchmark~\cite{geiger2012cvpr} measures the average precision at IoU=0.7.
In this section, we use JIoU as an extension of IoU by considering spatial uncertainty distribution, to evaluate the performance of several state-of-the-art deterministic and probabilistic object detectors. We report the detection performance in terms of the standard Average Precision (AP) or the Recall metrics from the Bird's Eye View on the KITTI dataset~\cite{geiger2012cvpr}, and show how JIoU provides additional information to analyze object detection compared to IoU.

\subsection{\textbf{Evaluating Deterministic Object Detectors}}\label{sec:experimental_result:jiou:dod}

\begin{figure*}[htpb]
	\centering
	\begin{minipage}{0.96\linewidth}
		\centering
		\includegraphics[width=1\linewidth]{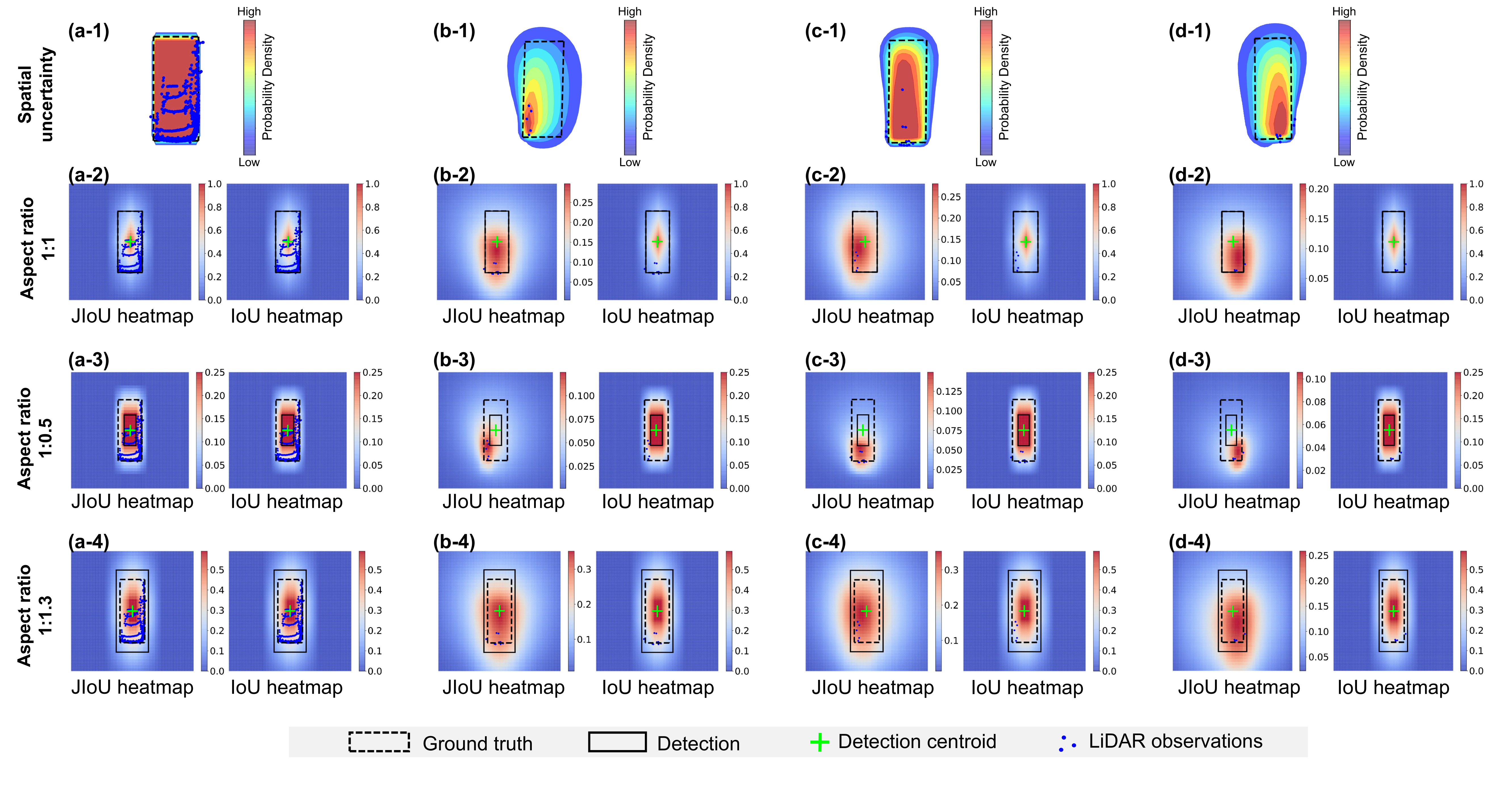}
		\caption{Synthetic analysis on how JIoU and IoU scores change with detections with different BBox locations and sizes. The JIoU and IoU heatmpas are generated by gradually shifting detections on the horizontal plane, and measuring the corresponding JIoU and IoU scores. First row: spatial uncertainty distributions of four grouth truth BBoxes. Second row: JIoU and IoU heatmaps of detections with their BBoxes equal to ground truths. Third row: length and width of detections are set $0.5\times$ of the ground truths. Fourth row: length and width of detections are set $1.3\times$ of the ground truths.} \label{fig:comparison_jiou_iou}
	\end{minipage}
\end{figure*}

First, we synthetically study the difference between JIoU and IoU. To do this, we fix the positions of ground truth bounding boxes (BBoxes), gradually shift their associated synthetic detections on the horizontal plane, and calculate the JIoU and IoU scores at each location. The JIoU scores are calculated between deterministic BBox predictions (which corresponds to uniform distribution in the JIoU calculation) and probabilistic ground truths (with inferred label uncertainty). Fig.~\ref{fig:comparison_jiou_iou} shows four examples. (a-1)-(d-1) demonstrate the spatial uncertainty distributions of ground truth BBoxes, (a-2)-(d-2) the heatmaps of JIoU and IoU scores when moving detections with the same shapes as BBoxes, and (a-3)-(d-3) the heatmaps of detections with their length and width setting to $0.5\times$ of the ground truths and (a-4)-(d-4) $1.3\times$ of the ground truths. We observe that with small label uncertainty (Fig.~\ref{fig:comparison_jiou_iou}(a-1)), the spatial uncertainty distribution of ground truth is uniformly distributed. In this case, JIoU behaves similar to IoU, as illustrated in (a-2)-(a-4). However, when the label uncertainty is high due to sparse LiDAR points or occlusion, cf. (b)-(d), JIoU shows very different behaviours from IoU. While IoU scores changes symmetrically, JIoU scores are biased towards to the regions with low label uncertainty in ground truth BBoxes, especially when the size of detection BBoxes is smaller than ground truths, cf. the third row of Fig.~\ref{fig:comparison_jiou_iou}. In other words, \textit{JIoU encourages detections which match the regions with low label uncertainty, and ignores those with high label uncertainty. In contrast, IoU only considers the geometric overlap ratio between BBoxes regardless of the distributions of LiDAR points.}

\begin{figure*}[!htpb]
	\centering
	\begin{minipage}{0.92\linewidth}
		\centering
		\includegraphics[width=1\linewidth]{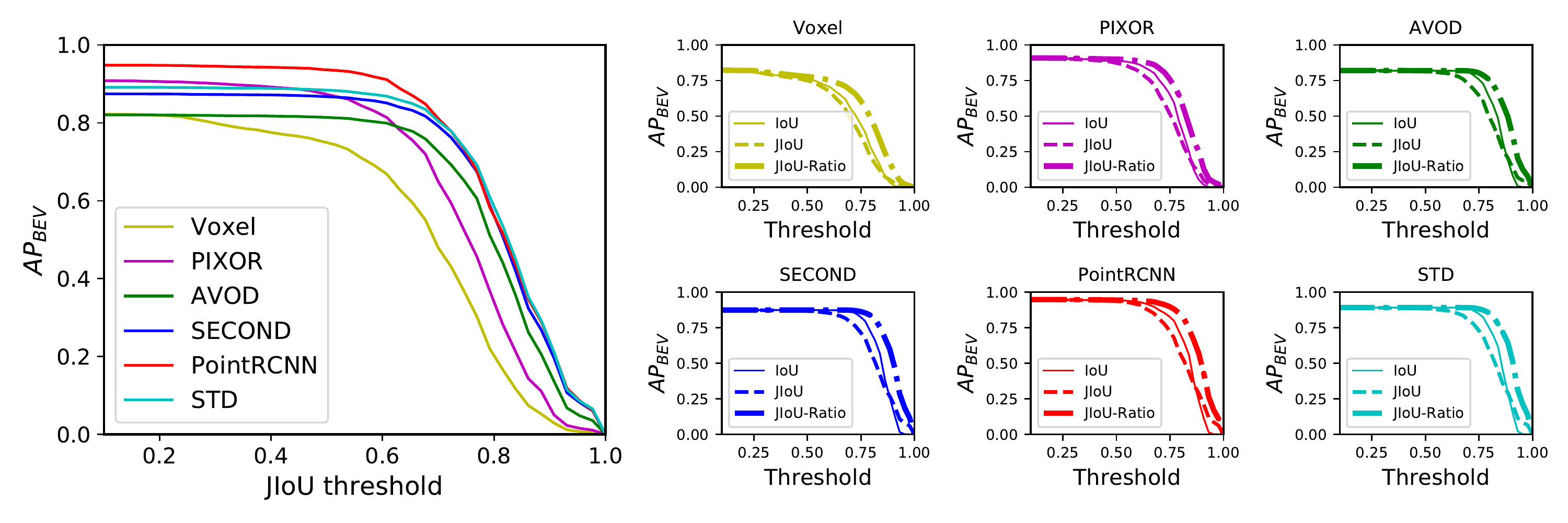}
        \caption{The evolution of average precision w.r.t. the increasing localization thresholds. The leftmost plot compares the average precision with JIoU threshold from six well-performed deterministic object detectors. The right six subplots compare the average precision with IoU, JIoU and JIoU-Ratio for each detector separately. }\label{fig:ap_jiou_all}
	\end{minipage}
\end{figure*}

Next, we use JIoU and IoU to compare among several state-of-the-art deterministic LiDAR-based object detection networks. Besides using JIoU scores in detection, we also calculate their ratios with the JIoU scores in ground truths (``JIoU-GT"), called ``JIoU-Ratio''. As introduced in Sec.~\ref{sec:uncertainty_justification:setup}, JIoU-GT reflects the uncertainty in ground truth labels. Therefore, JIoU-Ratio is a localization accuracy upper-bounded by the label uncertainty. A detection with low IoU or JIoU scores may still achieve high JIoU-Ratio, because its associated ground truth label is uncertain. In this case, the detector can be regarded to have sufficient detection performance considering the label difficulty. Therefore, JIoU-Ratio provides us extra information when evaluating the detection performance. Tab.~\ref{tab:detection_performance_jiou} reports the mean Average Precision (mAP) of six detectors averaged over localization thresholds (IoU, JIoU and JIoU-Ratio) from $0.5$ to $0.9$. Additionally, Fig.~\ref{fig:ap_jiou_all} shows the evolution of average precision w.r.t. the increasing localization thresholds. From the table and figures we have two observations. First, mAPs have the similar tendency among three localization metrics. Detectors which perform well in IoU thresholds also show superior results in JIoU or JIoU-Ratio (e.g. comparing STD and PIXOR). Second, mAPs in IoU are generally higher than those in JIoU but lower in JIoU-Ratio, because JIoU down-weights detections when their associated ground truths have high label uncertainty, which reduces recall rates. JIoU-Ratio removes this effect by judging the localization performance upper-bounded by the label uncertainty. 
\begin{table}[!tbp]
	\centering
	\caption{\small Comparison of LiDAR detectors in $\text{AP}_{BEV} (\%)$.}\label{tab:detection_performance_jiou}
	\resizebox{1\linewidth}{!}{\begin{tabular}{l|c c c}
			\Xhline{3\arrayrulewidth}
			\multirow{2}{*}{Methods} & \multicolumn{3}{c}{$\text{mAP}_{BEV} (\%)$ $\uparrow$ }\\ \cline{2-4}
			& IoU@0.5:0.9 & JIoU@0.5:0.9 & JIoU-ratio@0.5:0.9 \\ \hline 
			Voxel~\cite{zhou2017voxelnet} & $45.23$ & $41.63$ & $55.51$ \\
			PIXOR~\cite{yang2018pixor} & $58.67$ & $53.84$ & $67.54$ \\
			AVOD~\cite{ku2017joint} & $63.89$ & $59.18$ & $71.97$ \\
			SECOND~\cite{yan2018second} & $69.82$ & $65.10$ & $78.83$ \\
			PointRCNN~\cite{Shi_2019_CVPR} & $71.90$ & $68.35$ & $82.06$ \\
			STD~\cite{yang2019std} & $72.26$ & $66.95$ & $81.09$ \\
			\Xhline{3\arrayrulewidth}
	\end{tabular}}
\end{table}

\begin{figure*}[!htpb]
	\centering
	\begin{minipage}{0.92\linewidth}
		\centering
		\includegraphics[width=1\linewidth]{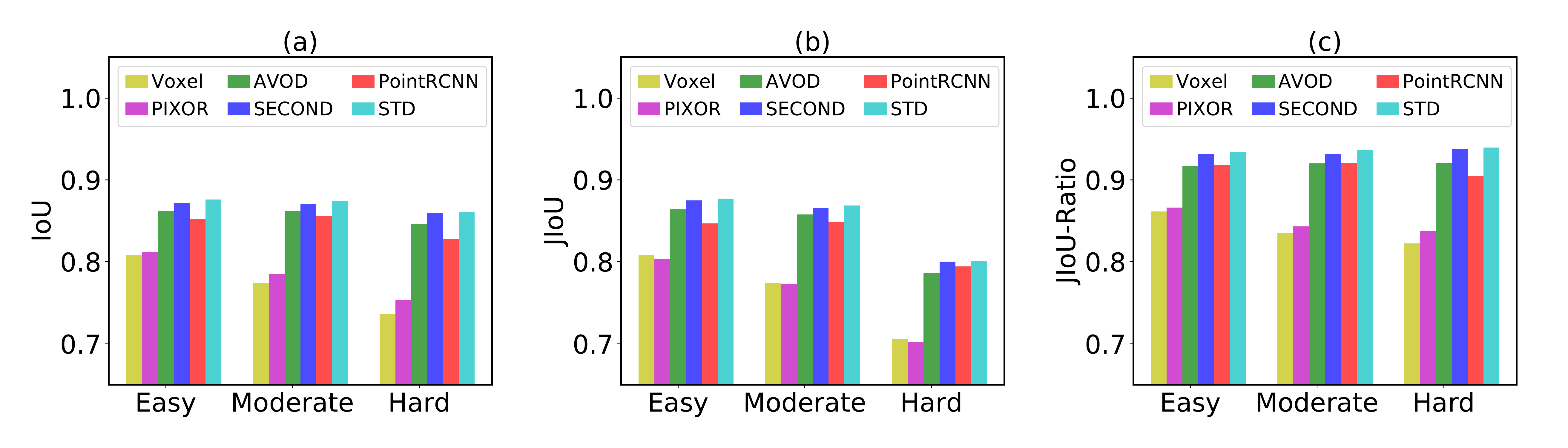}
		\caption{Comparison of the averaged (a) IoU, (b) JIoU, and (c) JIoU-Ratio for detections in the Easy, Moderate, and Hard settings defined by the KITTI dataset~\cite{geiger2012cvpr}.} \label{fig:comparison_jiou_difficulty}
	\end{minipage}
\end{figure*}

\begin{figure*}[!htpb]
	\centering
	\begin{minipage}{0.92\linewidth}
		\centering
		\includegraphics[width=1\linewidth]{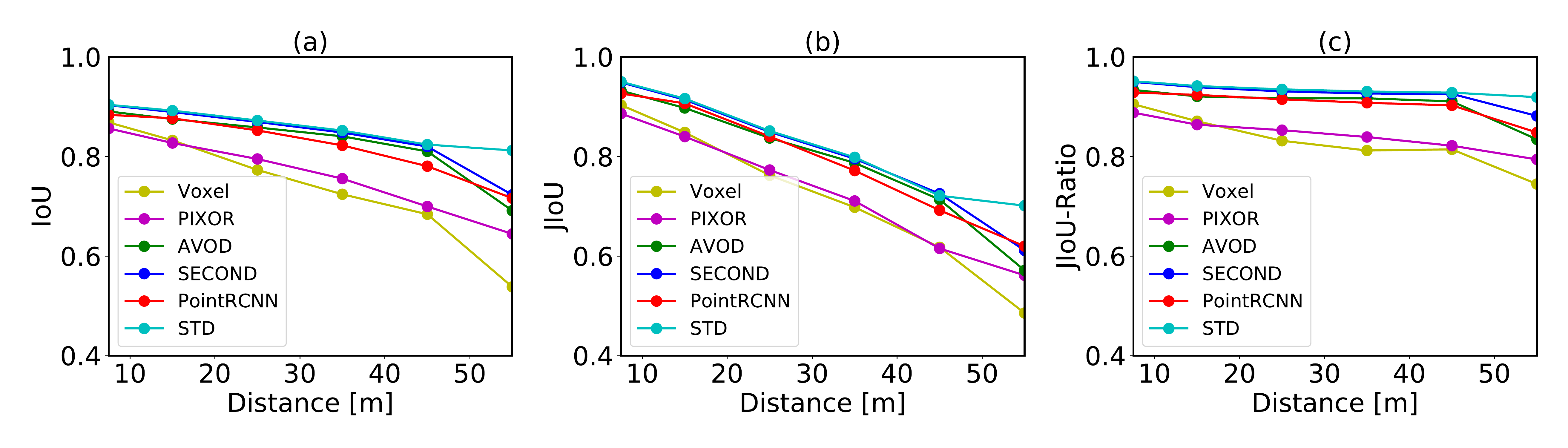}
		\caption{Comparison of the averaged (a) IoU, (b) JIoU, and (c) JIoU-Ratio for detections with increasing distance.} \label{fig:comparison_jiou_distance}
	\end{minipage}
\end{figure*}

Finally, we study how the localization accuracies in IoU, JIoU and JIoU-Ratio reflect the detection difficulty and distance. Fig.~\ref{fig:comparison_jiou_difficulty} shows the mean values of IoU, JIoU and JIoU-Ratio averaged over all detections in Easy, Moderate, and Hard settings defined in the KITTI dataset~\cite{geiger2012cvpr}. Fig.~\ref{fig:comparison_jiou_distance} shows their values with respect to detection distance. In general, we observe a slight decrease of averaged IoU with increasing detection difficulty (Fig.~\ref{fig:comparison_jiou_difficulty}(a)) or distance (Fig.~\ref{fig:comparison_jiou_distance}(a)) within all object detectors, indicating larger localization error in challenging cases. This trend is strengthened in averaged JIoU (Fig.~\ref{fig:comparison_jiou_difficulty}(b)), because JIoU incorporates spatial uncertainty distributions that are related to distance and detection difficulty. However, the trend is weakened in averaged JIoU-Ratio, which normalizes a JIoU score with the label uncertainty. Taking a closer look at each object detector, PIXOR and Voxel with inferior mAP (Tab.~\ref{tab:detection_performance_jiou}) also show smaller mean values and stronger performance drop in all three localization metrics. AVOD, SECOND and PointRCNN perform on par in localization accuracy, though AVOD clearly under-performs SECOND and PointRCNN in mAP. STD is the best detector with high localization accuracy. It remains constant JIoU-Ratio even in Hard setting or long range, showing its high detection performance. To conclude, JIoU together with its variant JIoU-Ratio can be used to in-depth analyze and compare object detectors in addition to IoU, especially when taking the detection difficulty and distance into account.

\subsection{\textbf{Evaluating Probabilistic Object Detectors}}
\begin{figure}[tpb]
	\centering
    \includegraphics[width=0.7\linewidth]{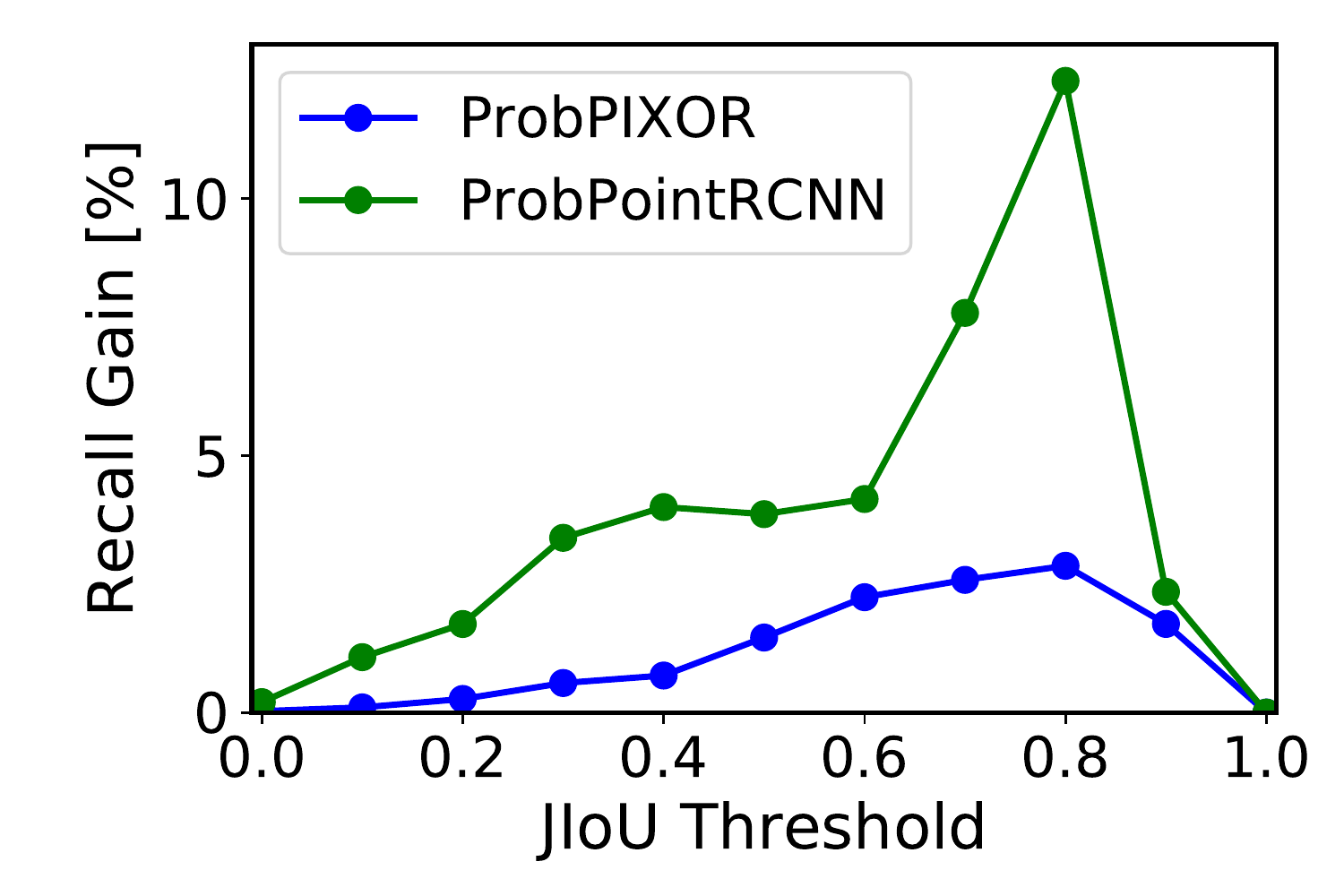}
	\caption{A study of JIoU on two probabilistic object detectors ProbPIXOR~\cite{feng2019can} and ProbPointRCNN~\cite{pan2020towards}. The gain of recall rates are measured by thresholding the JIoU scores from the detections with their predictive uncertainty estimates, compared to detections without uncertainty estimates.} \label{fig:jiou_recall_gain}
	\vspace{-10mm}
\end{figure}
Different from deterministic object detectors which only predict BBox regression variables, probabilistic object detectors additionally provide uncertainty estimates for those predictions (for a brief introduction cf. Sec.~\ref{subsec:related_works:uncertainty_estimation}). In this section, we use JIoU to quantitatively analyze how modelling uncertainties in BBoxes affects detection performance. To this end, we employ two state-of-the-art probabilistic object detectors based on LiDAR point clouds, namely, ProbPIXOR~\cite{feng2019can} and ProbPointRCNN~\cite{pan2020towards}. The ProbPIXOR network models uncertainties from the PIXOR~\cite{yang2018pixor} network, by assuming a univarate Gaussian distribution on each of the bouding box parameters encoded by the box centroid positions, width, length, and orientation (center-aligned BBox encoding). The ProbPointRCNN is based on PointRCNN~\cite{Shi_2019_CVPR}, and places a Laplacian distribution (i.e. the square root of a Gaussian distribution) on each of the eight corner positions in a 3D BBox (eight-points BBox encoding). We report the gain of recall rates, when thresholding the JIoU scores from the detections with their predictive uncertainty estimates, compared to detections without uncertainty estimates. In other words, we study the change of recall rates, purely from considering network's predictive uncertainties or not. Note that both networks will have zero recall gain, when using IoU scores for thresholding, because IoU remains constant regardless of uncertainty estimates. Fig.~\ref{fig:jiou_recall_gain} shows the experimental results. We observe consistent recall improvements in both networks when modelling uncertainties, indicating better detection performance. Both networks show largest recall gain at $0.8$ threshold, because the majority of detections lie at IoU values around $0.6-0.8$. Furthermore, the recall gain from ProbPointRCNN is larger than that from ProbPIXOR, because ProbPointRCNN estimates asymmetric uncertainty in each BBox corner, and thus captures more detailed uncertainties that reflect occlusion and truncation (as shown in \cite{pan2020towards}). However, ProbPIXOR only produces symmetric uncertainties due to its BBox encoding. To conclude, this experiment illustrates that our proposed JIoU scores can be used to compare probabilistic object detectors with different uncertainty estimates.
\section{\textbf{Using Label Uncertainty to Improve a Probabilistic Object Detector}}\label{sec:experimental_result:detection}
Finally, we leverage the proposed label uncertainty to train a probabilistic object detector and improve its detection performance during inference. In this section, we will treat each bounding box (BBox) regression variable separately. For simplicity, we use slightly different notation from Sec.~\ref{sec:methodology}. Here, $y$ is a regression variable instead of a vector, $x$ is a network's input LiDAR points instead of $X_{all}$ or $x_{1:K}$, and $q(\cdot)$ is an arbitrary probability distribution instead of the variational distribution. 

\subsection{\textbf{Methodology}}\label{sec:experimental_result:detection:methodology}
\subsubsection{\textbf{Training a Probabilistic Object Detector with the Kullback-Leibler Divergence (KLD) Loss}}
In recent years, many methods for probabilistic object detection have been proposed (a brief introduction cf. Sec.~\ref{subsec:related_works:uncertainty_estimation}). A prevalent method is to assume a certain probability distribution over the detector outputs, and learn the parameters for such a distribution by minimizing the negative log likelihood (NLL). For example, in~\cite{feng2019can} we build a state-of-the-art probabilistic object detector called ``ProbPIXOR'', which models uncertainty on a deterministic LiDAR-based object detector PIXOR~\cite{yang2018pixor}. The network encodes the BBox by its centroid positional offsets on the horizontal plane $\Delta x$, $\Delta y$, length and width in the log scale $\log(l)$, $\log(w)$, and orientation $\sin(r_y), \cos(r_y)$. Given an input sample of LiDAR point cloud $x$, ProbPIXOR assumes that each parameter $y$ in the BBox regression vector, i.e. $y\in \{\Delta x, \Delta y, \log(l), \log(w), \sin(r_y), \cos(r_y)\}$, follows a univariate Gaussian distribution, $q(y|x) = \mathcal{N}(y|\hat{y},\hat{\sigma}_q^2)$, with its mean and variance being directly predicted by the network output layers. The NLL for $q(y|x)$ results in an attenuated regression loss proposed by~\cite{kendall2017uncertainties}:
\begin{equation} \label{eq:attenuated_loss}
\mathcal{L}_{\text{NLL}} = \frac{\log \hat{\sigma}_q^2}{2} + \frac{(\overline{y}-\hat{y})^2}{2\hat{\sigma}_q^2}
\end{equation}
with $\overline{y}$ the label value of BBox provided by human annotators. 

One major problem of Eq.~\ref{eq:attenuated_loss} is that the object detector learns to predict the variance $\hat{\sigma}^2$ in an un-supervised manner, without any ground truth information. This may lead to unstable training process or suboptimal detection performance~\cite{wirges2019capturing}. He \textit{et al}~\cite{he2019bounding} and Meyer \textit{et al}~\cite{meyer2019learning} tackle this problem by assuming a simple prior knowledge of the BBox distribution, and minimizing its Kullback-Leibler Divergence (KLD) with the predictive probability. In this way, the network is regularized to predict probability close to the prior distribution. Since such prior knowledge is often related to the generation process of ground truth BBox parameters, it is also referred to as the label uncertainty, similar to the motivation in our work. For example, we can assume that a BBox regression variable $y$ is Gaussian distributed with $p(y|x) = \mathcal{N}(y|\overline{y}, \sigma_p^2)$, whose mean value is provided by human annotators, and its variance $\sigma_p^2$ indicates the strength of label uncertainty. To train ProbPIXOR, we minimize the KLD between $p$ and $q$ with $D_{KL}(p||q) = \int p(y|x)\log(\frac{p(y|x)}{q(y|x)})\text{d}y$, resulting in a closed-form KLD loss solution:
\begin{equation} \label{eq:kl-loss}
\mathcal{L}_{\text{KLD}} = \log\frac{\hat{\sigma}_q}{\sigma_p} + \frac{\sigma_p^2+(\overline{y}-\hat{y})^2}{2\hat{\sigma}_q^2}.
\end{equation}
The effectiveness of this KLD loss is largely dependent on the choice of $\sigma_p^2$. On the one hand, small $\sigma_p^2$ indicates that labels are accurate. In fact, when $\sigma_p^2\rightarrow0$, the label distribution becomes the Dirac-delta function~\cite{he2019bounding}, and the derivatives of Eq.~\ref{eq:kl-loss} degenerates to Eq.~\ref{eq:attenuated_loss}. On the other hand, large $\sigma_p^2$ indicates that labels are not trustful, and strongly regularizes predictive variances due to the $\sigma_p^2/(2\hat{\sigma}_q^2)$ term. Ideally, accurate samples are encouraged to train with low label uncertainty, and error-prone samples are penalized with large label uncertainty. However, both methods~\cite{he2019bounding,meyer2019learning} approximate the label uncertainty with simple heuristics, which do not fully reflect its behaviours: He \textit{et al.}~\cite{he2019bounding} ignore label noises by setting a Dirac-delta function on labels. Meyer \textit{et al.}~\cite{meyer2019learning} approximate label uncertainty by the intersection over union between a BBox and its convex hull of the aggregated LiDAR observations.

Our proposed label uncertainty has been shown to not only reflects complex environmental noises inherent in LiDAR perception, such as typical L-shape and occlusion, but also shows the quality of BBox labels in datasets (cf. Sec.~\ref{sec:uncertainty_justification}). Therefore, we combine the benefits of the KLD-loss~\cite{meyer2019learning} and our \textit{proposed} label uncertainty to train ProbPIXOR. 

\subsubsection{\textbf{Incorporating Label Uncertainty in the KLD Loss}}
We use our proposed label uncertainty to estimate $\sigma_p$. Recap that Sec.~\ref{sec:methodology:label_uncertainty} has introduced a variational distribution $q(y)$ to approximate the \textit{actual} distribution $p(y|x)$. The inferred full covariance matrix $\Sigma$ in Eq.~\ref{eq:VB_solution_covariance} provides correlations between regression variables parametrized by $\phi(y)$. For simplicity, however, we directly use the variances in this covariance matrix $\Sigma$ to represent the label uncertainty for Eq.~\ref{eq:kl-loss} while ignoring their correlations. We further use the error propagation technique~\cite{ku1966notes} to transform the variances from $\phi(y)$ to $y$. Intuitively, we can also approximate label uncertainty with ``num points'' and ``covx hull'' heuristics. The method ``num points'' scales the label uncertainty according to the number of LiDAR observations within a ground truth BBox, with the assumption that objects with increasingly sparse LiDAR points are more difficult to be labelled. The method ``covx hull'' proposed by~\cite{meyer2019learning} approximates the label uncertainty by measuring the IoU value between a BBox label and its convex hull of aggregated LiDAR points. In other words, this heuristic assumes that highly occluded objects tend to be error-prone.

\subsection{\textbf{Experimental Results}}\label{sec:experimental_result:detection:results}
We validate the effect of the proposed label uncertainty in training the ProbPIXOR network on detecting ``Car" objects in the KITTI dataset~\cite{geiger2012cvpr}. For comparison, we also train several ProbPIXOR networks with the label uncertainty estimated by the ``num points'' and ``covx hull'' heuristics, as well as some fixed values. All networks are trained with the SGD optimizer and the learning rate of $10^{-3}$ up to $140,000$ steps. We use the LiDAR point cloud within the range length $\times$ width $\times$ height = $[0,70]$m$\times[-40,40]$m$\times[0,2.5]$m, and do discretization with $\unit[0.1]{m}$ resolution. Similar to PIXOR~\cite{yang2018pixor}, we use global data augmentation by rotating and translating point clouds. Results are illustrated in Tab.~\ref{tab:detection_performance}. 

\begin{table}[!tbp]
	\centering
	\caption{\small Comparison of detection performance in $\text{AP}_{BEV} (\%)$ .}\label{tab:detection_performance}
	\resizebox{1\linewidth}{!}{\begin{tabular}{l|l l l}
			\Xhline{3\arrayrulewidth}
			\multirow{2}{*}{Methods} & \multicolumn{3}{c}{$\text{AP}_{BEV} (\%)$ for IoU@0.7 $\uparrow$ }\\ \cline{2-4}
			& Easy & Moderate & Hard \\ \hline 
			$\mathcal{L}_{\text{NLL}}$ (Baseline) & $88.60$ & $80.44$ & $78.74$ \\ \hline
			$\mathcal{L}_{\text{KLD}}$ (num points) & $90.74$ ($+2.14$) & $81.69$ ($+1.25$) & $79.35$ ($+0.61$) \\
			$\mathcal{L}_{\text{KLD}}$ (covx hull) & $90.05$ ($+1.45$) & $81.12$ ($+0.68$) & $78.84$ ($+0.10$)  \\
			\rowcolor{Gray} $\mathcal{L}_{\text{KLD}}$ (\textbf{Ours}) & $\mathbf{92.22}$ ($\mathbf{+3.62}$) & $\mathbf{82.03}$ ($\mathbf{+1.59}$) & $79.16$ ($+0.42$) \\ \hline
			$\mathcal{L}_{\text{KLD}}$ ($\sigma_p^2=10^{-4}$) & $90.13$ ($+1.53$) & $81.44$ ($+1.00$) & $78.03$ ($-0.71$)  \\ 
			$\mathcal{L}_{\text{KLD}}$ ($\sigma_p^2=10^{-3}$) & $89.82$ ($+1.22$) & $81.79$ ($+1.35$) & $\mathbf{79.54}$ ($\mathbf{+0.80}$)  \\
			$\mathcal{L}_{\text{KLD}}$ ($\sigma_p^2=10^{-2}$) & $89.09$ ($+0.49$) & $81.33$ ($+0.89$) & $79.17$ ($+0.43$)  \\
			$\mathcal{L}_{\text{KLD}}$ ($\sigma_p^2=10^{-1}$) & $83.82$ ($-4.78$) & $79.25$ ($-1.19$) & $77.69$ ($-1.05$)  \\ 
			$\mathcal{L}_{\text{KLD}}$ ($\sigma_p^2=10^{0}$) & $69.17$ ($-19.43$) & $72.56$ ($-7.88$) & $71.43$ ($-7.31$)  \\ 
			$\mathcal{L}_{\text{KLD}}$ ($\sigma_p^2=10^{1}$) & $14.99$ ($-73.61$) & $18.23$ ($-62.21$) & $16.18$ ($-62.56$)  \\ 
			\Xhline{3\arrayrulewidth}
	\end{tabular}}
\end{table}
First, we build a baseline probabilistic object detector called ``ProbPIXOR + $\mathcal{L}_{\text{NLL}}$''~\cite{feng2019can}, which learns to predict uncertainty by minimizing the negative log likelihood (NLL), following Eq.~\ref{eq:attenuated_loss}. It reaches better or on par detection performance compared to the original PIXOR network~\cite{yang2018pixor}. Next, we train ProbPIXOR by the KLD-loss following Eq.~\ref{eq:kl-loss}, and compare among different methods differentiating from how they define label uncertainty (characterized by the variance $\sigma_p^2$). We compare among the uncertainty extracted based on two simple heuristics ``num points'' and ``covx hull'', and the uncertainty from the proposed generative model (``Ours''). From the table we observe that the networks based on the KLD-loss improve the detection performance compared to the baseline model, showing the benefits of the uncertainty regularization effect in the KLD-loss. More importantly, our method (``Ours'') provides the largest performance gain especially at the ``Easy'' setting by $3.6\%$ AP. This indicates that our method produces better label uncertainty than the simple heuristics that rely on the number of points or the convex hull.

As discussed in the previous section, Sec.~\ref{sec:experimental_result:detection:methodology}, the choice of label uncertainty has a big impact on the training procedure. In Tab.~\ref{tab:detection_performance}, we additionally study how networks perform when being trained with increasing \textit{fixed} label uncertainty for all training samples. We observe an improvement of $AP_{BEV}$ scores compared to the baseline method when setting label uncertainty $\sigma_p^2<10^{-2}$. However, the detection performance drops significantly at larger label uncertainty $\sigma_p^2>10^{-2}$. This result suggests the necessity of setting a proper range of label uncertainty. Under-confident labels (or too large variances in other words) decrease detection accuracy.

In conclusion, we can leverage the proposed label uncertainty to regularize the training of a probabilistic LiDAR-based object detector with the help of the KLD loss, and to improve its detection accuracy.

\section{\textbf{Conclusion}} \label{sec:conclusion}
We have presented our methods which (1). infer the uncertainty inherent in bounding box labels using LiDAR points, (2). visualize probabilistic bounding boxes by the spatial uncertainty distributions, (3). apply label uncertainty to evaluate object detectors via the novel Jaccard IoU (JIoU) metric, and (4). apply label uncertainty to improve the training of a probabilistic object detector. Comprehensive experiments on two public datasets (KITTI and Waymo) and simulation data verified the proposed methods. 

There are several ways to extend the proposed methods. First, this work only models label uncertainty for bounding boxes of ``car'' objects. The label uncertainty could be extended to other classes as well, such as ``pedestrian'' and ``cyclist'', using more sophisticated LiDAR point registration models. Second, the proposed model only captures the \textit{local} measurement noise for each object separately. An interesting future work is to consider the \textit{global} measurement noise, by taking the positions of multiple objects into account (e.g. a sequence of parked cars by the side of a road are difficult to be annotated due to occlusion). Finally, this work models label uncertainty by the covariance matrix from the inferred posterior distribution of bounding box parameters. The inferred mean values are different from human labels, which are however ignored in this work, as we assume that human labels are unbiased estimators of bounding box parameters (cf. the third assumption in Sec.~\ref{sec:methodology}). To study the inferred mean values of the posterior distribution, it is necessary to construct the ``ground truths'' of human labelling errors, e.g. by asking several human annotators to label objects with known locations and extents. We leave it as an interesting future work.

\section*{\textbf{Acknowledgment}}
This work was supported by a fellowship within the IFI program of the German Academic Exchange Service (DAAD). We also thank National Instrument for providing the car simulator, and Hujie Pan for supporting experiments and insightful discussions. 

\bibliographystyle{IEEEtran}{}
\bibliography{IEEEabrv,bibliography}

\begin{thebibliography}{10}
\providecommand{\url}[1]{#1}
\csname url@samestyle\endcsname
\providecommand{\newblock}{\relax}
\providecommand{\bibinfo}[2]{#2}
\providecommand{\BIBentrySTDinterwordspacing}{\spaceskip=0pt\relax}
\providecommand{\BIBentryALTinterwordstretchfactor}{4}
\providecommand{\BIBentryALTinterwordspacing}{\spaceskip=\fontdimen2\font plus
\BIBentryALTinterwordstretchfactor\fontdimen3\font minus
  \fontdimen4\font\relax}
\providecommand{\BIBforeignlanguage}[2]{{%
\expandafter\ifx\csname l@#1\endcsname\relax
\typeout{** WARNING: IEEEtran.bst: No hyphenation pattern has been}%
\typeout{** loaded for the language `#1'. Using the pattern for}%
\typeout{** the default language instead.}%
\else
\language=\csname l@#1\endcsname
\fi
#2}}
\providecommand{\BIBdecl}{\relax}
\BIBdecl

\bibitem{geiger2012cvpr}
A.~Geiger, P.~Lenz, and R.~Urtasun, ``Are we ready for autonomous driving? the
  {KITTI} vision benchmark suite,'' in \emph{IEEE Conference on Computer Vision
  and Pattern Recognition (CVPR)}, 2012.

\bibitem{sun2019scalability}
P.~Sun, H.~Kretzschmar, X.~Dotiwalla, A.~Chouard, V.~Patnaik, P.~Tsui, J.~Guo,
  Y.~Zhou, Y.~Chai, B.~Caine, V.~Vasudevan, W.~Han, J.~Ngiam, H.~Zhao,
  A.~Timofeev, S.~Ettinger, M.~Krivokon, A.~Gao, A.~Joshi, Y.~Zhang, J.~Shlens,
  Z.~Chen, and D.~Anguelov, ``Scalability in perception for autonomous driving:
  Waymo open dataset,'' in \emph{IEEE Conference on Computer Vision and Pattern
  Recognition (CVPR)}, 2020, pp. 2446--2454.

\bibitem{janai2017computer}
J.~Janai, F.~G{\"u}ney, A.~Behl, A.~Geiger \emph{et~al.}, ``Computer vision for
  autonomous vehicles: Problems, datasets and state of the art,''
  \emph{Foundations and Trends in Computer Graphics and Vision}, vol.~12, no.
  1--3, pp. 1--308, 2020.

\bibitem{sukhbaatar2014training}
S.~Sukhbaatar, J.~Bruna, M.~Paluri, L.~Bourdev, and R.~Fergus, ``Training
  convolutional networks with noisy labels,'' in \emph{International Conference
  on Learning Representations Workshop (ICLRW)}, 2014.

\bibitem{lawrence2001estimating}
N.~D. Lawrence and B.~Sch{\"o}lkopf, ``Estimating a kernel fisher discriminant
  in the presence of label noise,'' in \emph{International Conference on
  Machine Learning (ICML)}, vol.~1, 2001, pp. 306--313.

\bibitem{xiao2015learning}
T.~Xiao, T.~Xia, Y.~Yang, C.~Huang, and X.~Wang, ``Learning from massive noisy
  labeled data for image classification,'' in \emph{IEEE Conference on Computer
  Vision and Pattern Recognition (CVPR)}, 2015, pp. 2691--2699.

\bibitem{vahdat2017toward}
A.~Vahdat, ``Toward robustness against label noise in training deep
  discriminative neural networks,'' in \emph{Advances in Neural Information
  Processing Systems (NeurIPS)}, 2017, pp. 5596--5605.

\bibitem{haase2019estimate}
C.~{Haase-Sch\"utz}, H.~{Hertlein}, and W.~{Wiesbeck}, ``Estimating labeling
  quality with deep object detectors,'' in \emph{IEEE Intelligent Vehicles
  Symposium (IV)}, June 2019, pp. 33--38.

\bibitem{everingham2010pascal}
M.~Everingham, L.~Van~Gool, C.~K. Williams, J.~Winn, and A.~Zisserman, ``The
  pascal visual object classes ({VOC}) challenge,'' \emph{International Journal
  of Computer Vision (IJCV)}, vol.~88, no.~2, pp. 303--338, 2010.

\bibitem{oksuz2018localization}
K.~Oksuz, B.~Can~Cam, E.~Akbas, and S.~Kalkan, ``Localization recall precision
  ({LRP}): A new performance metric for object detection,'' in \emph{European
  Conference on Computer Vision (ECCV)}, 2018, pp. 504--519.

\bibitem{feng2018towards}
D.~Feng, L.~Rosenbaum, and K.~Dietmayer, ``Towards safe autonomous driving:
  {Capture} uncertainty in the deep neural network for {Lidar} 3d vehicle
  detection,'' in \emph{IEEE International Conference Intelligent
  Transportation System (ITSC)}, Nov. 2018, pp. 3266--3273.

\bibitem{miller2017dropout}
D.~Miller, L.~Nicholson, F.~Dayoub, and N.~S{\"u}nderhauf, ``Dropout sampling
  for robust object detection in open-set conditions,'' in \emph{IEEE
  International Conference on Robotics and Automation (ICRA)}, 2018.

\bibitem{harakeh2019bayesod}
A.~Harakeh, M.~Smart, and S.~Waslander, ``{BayesOD}: A bayesian approach for
  uncertainty estimation in deep object detectors,'' in \emph{IEEE
  International Conference on Robotics and Automation (ICRA)}, 2020, pp.
  87--93.

\bibitem{miller2019evaluating}
D.~Miller, F.~Dayoub, M.~Milford, and N.~S{\"u}nderhauf, ``Evaluating merging
  strategies for sampling-based uncertainty techniques in object detection,''
  in \emph{IEEE International Conference on Robotics and Automation (ICRA)},
  2019, pp. 2348--2354.

\bibitem{hall2018probabilistic}
D.~Hall, F.~Dayoub, J.~Skinner, H.~Zhang, D.~Miller, P.~Corke, G.~Carneiro,
  A.~Angelova, and N.~S{\"u}nderhauf, ``Probabilistic object detection:
  Definition and evaluation,'' in \emph{IEEE Winter Conference on Applications
  of Computer Vision (WACV)}, 2020, pp. 1031--1040.

\bibitem{frenay2013classification}
B.~Fr{\'e}nay and M.~Verleysen, ``Classification in the presence of label
  noise: a survey,'' \emph{IEEE Transactions on Neural Networks and Learning
  Systems}, vol.~25, no.~5, pp. 845--869, 2013.

\bibitem{algan2019image}
G.~Algan and I.~Ulusoy, ``Image classification with deep learning in the
  presence of noisy labels: A survey,'' \emph{arXiv preprint arXiv:1912.05170},
  2019.

\bibitem{bishop2006pattern}
C.~M. Bishop, \emph{Pattern recognition and machine learning}.\hskip 1em plus
  0.5em minus 0.4em\relax Springer, 2006.

\bibitem{meyer2019learning}
G.~P. Meyer and N.~Thakurdesai, ``Learning an uncertainty-aware object detector
  for autonomous driving,'' \emph{arXiv preprint arXiv:1910.11375}, 2019.

\bibitem{meyer2019huber}
G.~P. Meyer, ``An alternative probabilistic interpretation of the huber loss,''
  \emph{arXiv preprint arXiv:1911.02088}, 2019.

\bibitem{granstrom2011tracking}
K.~Granstr{\"o}m, C.~Lundquist, and U.~Orguner, ``Tracking rectangular and
  elliptical extended targets using laser measurements,'' in \emph{{IEEE}
  International Conference on Information Fusion (FUSION)}, 2011, pp. 1--8.

\bibitem{scheel2018tracking}
A.~Scheel and K.~Dietmayer, ``Tracking multiple vehicles using a variational
  radar model,'' \emph{IEEE Transactions on Intelligent Transportation Systems
  (TITS)}, vol.~20, no.~10, pp. 3721--3736, 2018.

\bibitem{hirscher2016tracking}
T.~{Hirscher}, A.~{Scheel}, S.~{Reuter}, and K.~{Dietmayer}, ``Multiple
  extended object tracking using gaussian processes,'' in \emph{{IEEE}
  International Conference on Information Fusion (FUSION)}, July 2016, pp.
  868--875.

\bibitem{rezatofighi2019generalized}
H.~Rezatofighi, N.~Tsoi, J.~Gwak, A.~Sadeghian, I.~Reid, and S.~Savarese,
  ``Generalized intersection over union: {A} metric and a loss for bounding box
  regression,'' in \emph{IEEE Conference on Computer Vision and Pattern
  Recognition (CVPR)}, 2019, pp. 658--666.

\bibitem{zhou2019iou}
D.~Zhou, J.~Fang, X.~Song, C.~Guan, J.~Yin, Y.~Dai, and R.~Yang, ``{IOU} loss
  for 2d/3d object detection,'' in \emph{International Conference on 3D Vision
  (3DV)}, 2019, pp. 85--94.

\bibitem{zheng2020distance}
Z.~Zheng, P.~Wang, W.~Liu, J.~Li, R.~Ye, and D.~Ren, ``Distance-{IoU} loss:
  Faster and better learning for bounding box regression,'' in \emph{The AAAI
  Conference of Artificial Intelligence}, 2020.

\bibitem{jiang2018acquisition}
B.~Jiang, R.~Luo, J.~Mao, T.~Xiao, and Y.~Jiang, ``Acquisition of localization
  confidence for accurate object detection,'' in \emph{European Conference on
  Computer Vision (ECCV)}, 2018, pp. 784--799.

\bibitem{he2020sassd}
C.~He, H.~Zeng, J.~Huang, X.-S. Hua, and L.~Zhang, ``Structure aware
  single-stage 3d object detection from point cloud,'' in \emph{IEEE Conference
  on Computer Vision and Pattern Recognition (CVPR)}.

\bibitem{lin2014microsoft}
T.-Y. Lin, M.~Maire, S.~Belongie, J.~Hays, P.~Perona, D.~Ramanan,
  P.~Doll{\'a}r, and C.~L. Zitnick, ``{Microsoft COCO}: Common objects in
  context,'' in \emph{European Conference on Computer Vision (ECCV)}, 2014, pp.
  740--755.

\bibitem{caesar2019nuscenes}
H.~Caesar, V.~Bankiti, A.~H. Lang, S.~Vora, V.~E. Liong, Q.~Xu, A.~Krishnan,
  Y.~Pan, G.~Baldan, and O.~Beijbom, ``nu{S}cenes: A multimodal dataset for
  autonomous driving,'' in \emph{IEEE Conference on Computer Vision and Pattern
  Recognition (CVPR)}, 2020, pp. 11\,621--11\,631.

\bibitem{wirges2019capturing}
S.~Wirges, M.~Reith-Braun, M.~Lauer, and C.~Stiller, ``Capturing object
  detection uncertainty in multi-layer grid maps,'' in \emph{IEEE Intelligent
  Vehicles Symposium (IV)}, 2019, pp. 1520--1526.

\bibitem{Gal2016Uncertainty}
Y.~Gal, ``Uncertainty in deep learning,'' Ph.D. dissertation, University of
  Cambridge, 2016.

\bibitem{lakshminarayanan2017simple}
B.~Lakshminarayanan, A.~Pritzel, and C.~Blundell, ``Simple and scalable
  predictive uncertainty estimation using deep ensembles,'' in \emph{Advances
  in Neural Information Processing Systems (NeurIPS)}, 2017, pp. 6402--6413.

\bibitem{feng2018leveraging}
D.~Feng, L.~Rosenbaum, F.~Timm, and K.~Dietmayer, ``Leveraging heteroscedastic
  aleatoric uncertainties for robust real-time {LiDAR} 3d object detection,''
  in \emph{IEEE Intelligent Vehicles Symposium (IV)}, 2019.

\bibitem{feng2019can}
D.~Feng, L.~Rosenbaum, C.~Gl{\"a}ser, F.~Timm, and K.~Dietmayer, ``Can we trust
  you? {On} calibration of a probabilistic object detector for autonomous
  driving,'' in \emph{IEEE/RSJ International Conference on Intelligent Robots
  and Systems Workshops (IROSW)}, 2019.

\bibitem{he2019bounding}
Y.~He, C.~Zhu, J.~Wang, M.~Savvides, and X.~Zhang, ``Bounding box regression
  with uncertainty for accurate object detection,'' in \emph{IEEE Conference on
  Computer Vision and Pattern Recognition (CVPR)}, 2019, pp. 2888--2897.

\bibitem{meyer2019lasernet}
G.~P. Meyer, A.~Laddha, E.~Kee, C.~Vallespi-Gonzalez, and C.~K. Wellington,
  ``Lasernet: {An} efficient probabilistic 3d object detector for autonomous
  driving,'' in \emph{IEEE Conference on Computer Vision and Pattern
  Recognition (CVPR)}, 2019, pp. 12\,677--12\,686.

\bibitem{choi2019gaussian}
J.~Choi, D.~Chun, H.~Kim, and H.-J. Lee, ``{Gaussian YOLOv3}: An accurate and
  fast object detector using localization uncertainty for autonomous driving,''
  in \emph{IEEE International Conference on Computer Vision (ICCV)}, October
  2019.

\bibitem{le2018uncertainty}
M.~T. Le, F.~Diehl, T.~Brunner, and A.~Knol, ``Uncertainty estimation for deep
  neural object detectors in safety-critical applications,'' in \emph{IEEE
  International Conference Intelligent Transportation System (ITSC)}, 2018, pp.
  3873--3878.

\bibitem{pan2020towards}
H.~Pan, Z.~Wang, W.~Zhan, and M.~Tomizuka, ``Towards better performance and
  more explainable uncertainty for 3d object detection of autonomous
  vehicles,'' in \emph{IEEE International Conference Intelligent Transportation
  System (ITSC)}, 2020.

\bibitem{dong2020probabilistic}
X.~Dong, P.~Wang, P.~Zhang, and L.~Liu, ``Probabilistic oriented object
  detection in automotive radar,'' in \emph{IEEE Conference on Computer Vision
  and Pattern Recognition Workshops (CVPRW)}, 2020, pp. 102--103.

\bibitem{feng2019deep}
D.~Feng, X.~Wei, L.~Rosenbaum, A.~Maki, and K.~Dietmayer, ``Deep active
  learning for efficient training of a lidar 3d object detector,'' in
  \emph{IEEE Intelligent Vehicles Symposium (IV)}, 2019.

\bibitem{guo2017calibration}
C.~Guo, G.~Pleiss, Y.~Sun, and K.~Q. Weinberger, ``On calibration of modern
  neural networks,'' in \emph{International Conference on Machine Learning
  (ICML)}, 2017, pp. 1321--1330.

\bibitem{kuleshov2018accurate}
V.~Kuleshov, N.~Fenner, and S.~Ermon, ``Accurate uncertainties for deep
  learning using calibrated regression,'' in \emph{International Conference on
  Machine Learning (ICML)}, 2018.

\bibitem{fang2018simulating}
J.~Fang, D.~Zhou, F.~Yan, T.~Zhao, F.~Zhang, Y.~Ma, L.~Wang, and R.~Yang,
  ``Augmented lidar simulator for autonomous driving,'' \emph{IEEE Robotics and
  Automation Letters}, vol.~5, no.~2, pp. 1931--1938, 2020.

\bibitem{blei2017variational}
D.~M. Blei, A.~Kucukelbir, and J.~D. McAuliffe, ``Variational inference: {A}
  review for statisticians,'' \emph{Journal of the American Statistical
  Association}, vol. 112, no. 518, pp. 859--877, 2017.

\bibitem{moulton2018maximally}
R.~Moulton and Y.~Jiang, ``Maximally consistent sampling and the jaccard index
  of probability distributions,'' in \emph{{IEEE} International Conference on
  Data Mining Workshop (ICDMW)}, 2018.

\bibitem{yang2018pixor}
B.~Yang, W.~Luo, and R.~Urtasun, ``Pixor: {Real-time} 3d object detection from
  point clouds,'' in \emph{IEEE Conference on Computer Vision and Pattern
  Recognition (CVPR)}, 2018, pp. 7652--7660.

\bibitem{zhang2017efficient}
X.~Zhang, W.~Xu, C.~Dong, and J.~M. Dolan, ``Efficient {L-shape} fitting for
  vehicle detection using laser scanners,'' in \emph{IEEE Intelligent Vehicles
  Symposium (IV)}, 2017, pp. 54--59.

\bibitem{monodrive}
``monodrive:autonomous vehicle simulator,'' \url{https://www.monodrive.io/},
  2020.

\bibitem{zhou2017voxelnet}
Y.~Zhou and O.~Tuzel, ``{VoxelNet}: {E}nd-to-end learning for point cloud based
  3d object detection,'' in \emph{IEEE Conference on Computer Vision and
  Pattern Recognition (CVPR)}, 2018.

\bibitem{ku2017joint}
J.~Ku, M.~Mozifian, J.~Lee, A.~Harakeh, and S.~Waslander, ``Joint 3d proposal
  generation and object detection from view aggregation,'' in \emph{IEEE/RSJ
  International Conference on Intelligent Robots and Systems (IROS)}, Oct.
  2018, pp. 1--8.

\bibitem{yan2018second}
Y.~Yan, Y.~Mao, and B.~Li, ``{SECOND}: {S}parsely embedded convolutional
  detection,'' \emph{Sensors}, vol.~18, no.~10, p. 3337, 2018.

\bibitem{Shi_2019_CVPR}
S.~Shi, X.~Wang, and H.~Li, ``{PointRCNN}: 3d object proposal generation and
  detection from point cloud,'' in \emph{IEEE Conference on Computer Vision and
  Pattern Recognition (CVPR)}, June 2019.

\bibitem{yang2019std}
Z.~Yang, Y.~Sun, S.~Liu, X.~Shen, and J.~Jia, ``{STD}: {S}parse-to-dense 3d
  object detector for point cloud,'' in \emph{IEEE Conference on Computer
  Vision and Pattern Recognition (CVPR)}, 2019, pp. 1951--1960.

\bibitem{kendall2017uncertainties}
A.~Kendall and Y.~Gal, ``What uncertainties do we need in {Bayesian} deep
  learning for computer vision?'' in \emph{Advances in Neural Information
  Processing Systems (NeurIPS)}, 2017, pp. 5574--5584.

\bibitem{ku1966notes}
H.~H. Ku \emph{et~al.}, ``Notes on the use of propagation of error formulas,''
  \emph{Journal of Research of the National Bureau of Standards}, vol.~70,
  no.~4, pp. 263--273, 1966.

\end{thebibliography}

\begin{IEEEbiography}[{\includegraphics[width=1in,height=1.25in,clip,keepaspectratio]{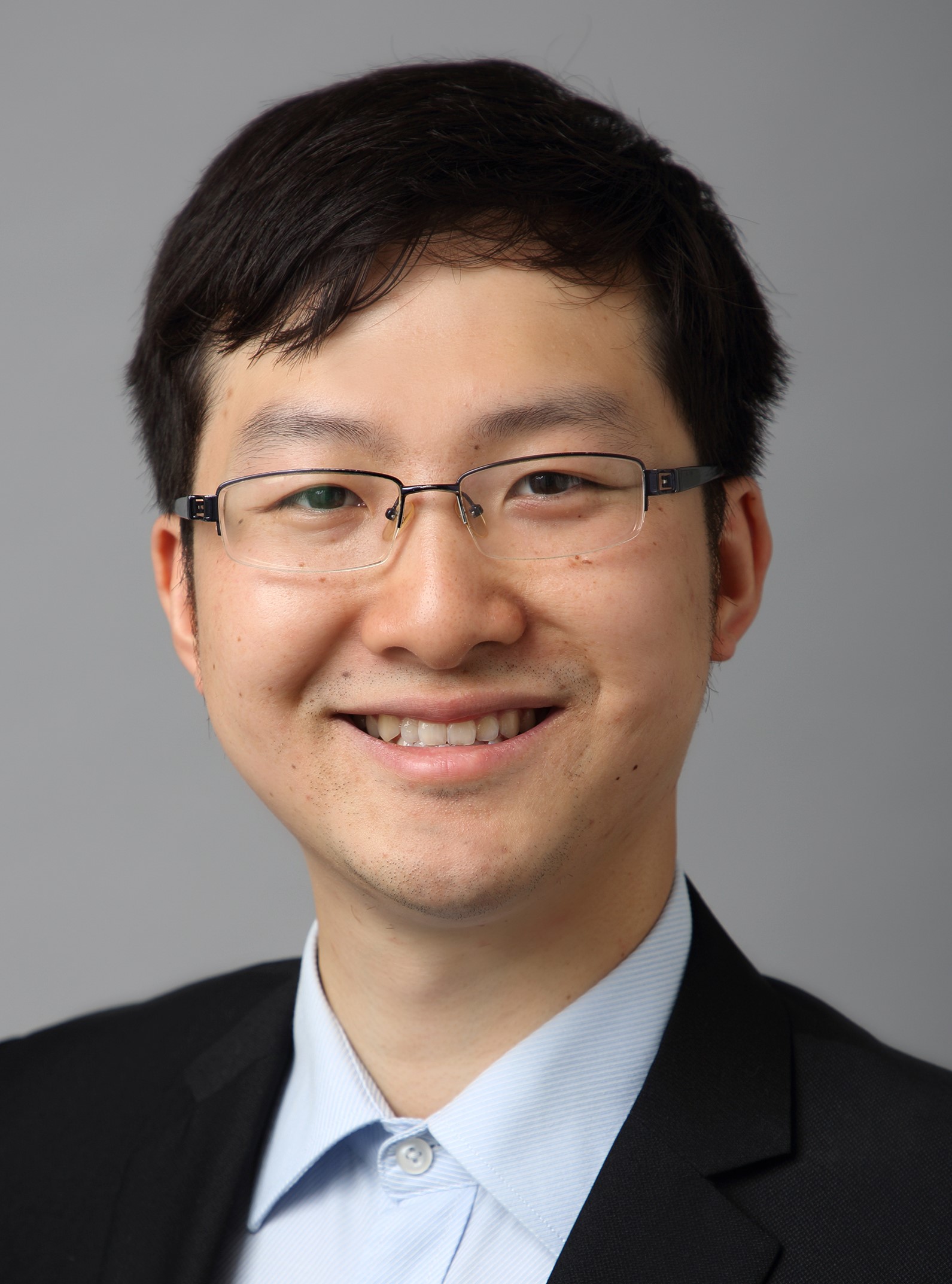}}]{Di Feng}
(Member, IEEE) is currently a visiting researcher in Mechanical Systems Control Laboratory at UC Berkeley. He did his doctoral degree in the autonomous driving group, Corporate Research, Robert Bosch GmbH, Stuttgart, in cooperation with the Ulm University. Prior joining Bosch, He finished his master's degree with distinction in electrical and computer engineering at the Technical University of Munich. During his studies, he was granted the opportunity to work in several teams with reputable companies and research institutes such as BMW AG, German Aerospace Center (DLR), and Institute for Cognitive Systems (ICS) at Technical University of Munich. He received the bachelor's degree in mechatronics with honor from Tongji University in 2014. His current research is centered on robust multi-modal perception using deep learning approaches for autonomous driving. He is also interested in general topics related to perception and learning in robotic and cognitive systems.
\end{IEEEbiography}
\vspace{-3mm}
\begin{IEEEbiography}
[{\includegraphics[width=1in,height=1.25in,clip,keepaspectratio]{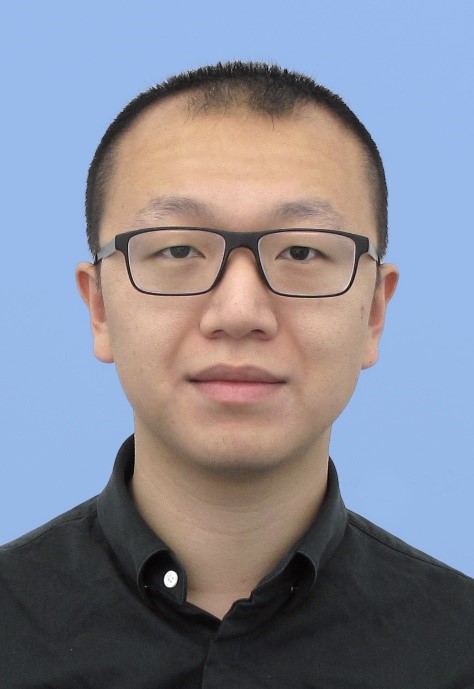}}]{Zining Wang} received the B.S. degree of engineering mechanics from Tsinghua University, Beijing, China, in 2015. He is now pursuing the PhD degree in mechanical engineering at University of California, Berkeley, CA, USA. His current research interests include perception with deep learning, sensor fusion and robust perception for autonomous driving. He is also interested in active calibration and adaptive control for industrial manipulators.
\end{IEEEbiography}
\vspace{-5mm}
\begin{IEEEbiography}
[{\includegraphics[width=1in,height=1.25in,clip,keepaspectratio]{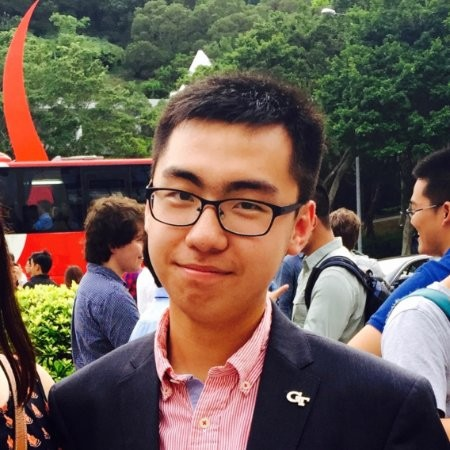}}]{Yiyang Zhou} graduated from the Georgia Institute of Technology with a B.S. degree in Mechanical Engineering in 2018, and he is now pursing a PhD degree in Controls at the University of California, Berkeley. His current research focuses at the front-end of autonomous driving-related algorithms, including hardware calibration, mapping, localization, and perception. 
\end{IEEEbiography}
\vspace{-5mm}
\begin{IEEEbiography}
[{\includegraphics[width=1in,height=1.25in,clip,keepaspectratio]{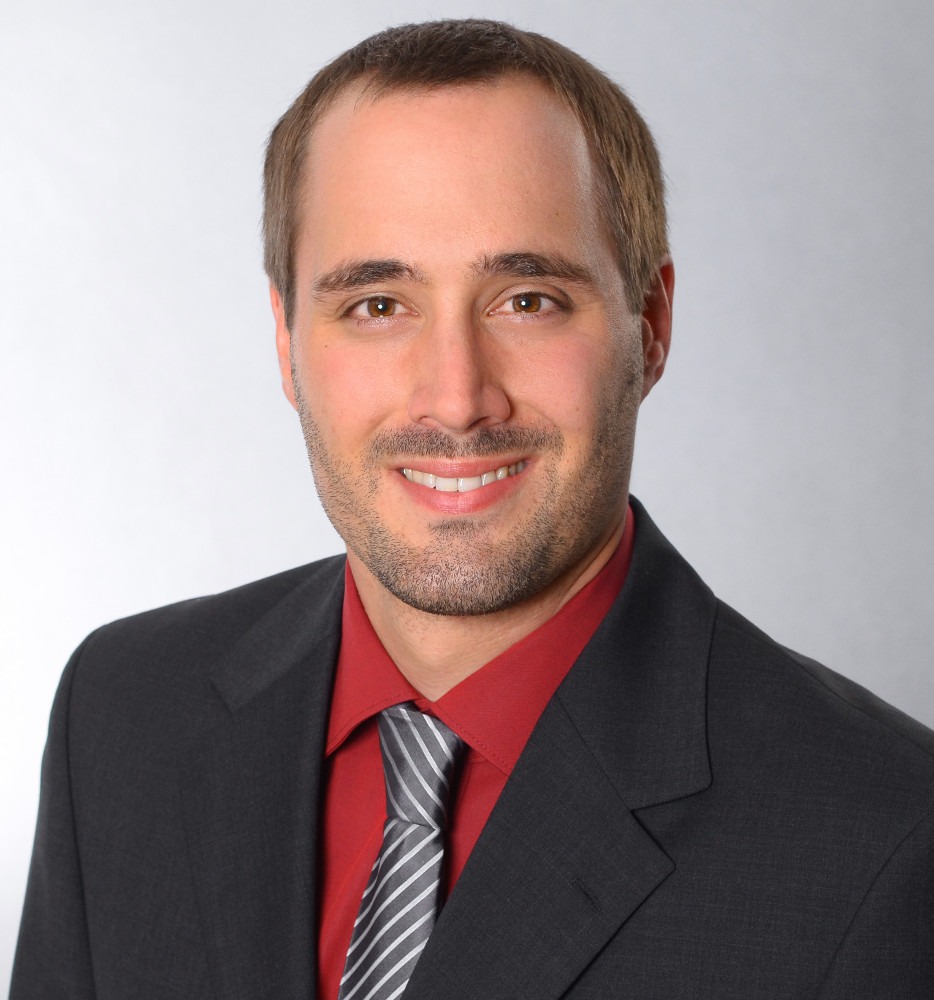}}]{Lars Rosenbaum} received his Dipl.-Inf. (M.S.) and the Dr. rer. nat. (Ph.D.) degrees in bioinformatics from the University of Tuebingen, Germany, in 2009 and 2013, respectively. During this time he was working on machine learning approaches for computer-aided molecular drug design and analysis of metabolomics data. In 2014, he joined ITK Engineering in Marburg, Germany, working on driver assistance systems. Since 2016, he is a research engineer at Corporate Research, Robert Bosch GmbH in Renningen, Germany, where he is currently doing research on machine learning algorithms in the area of perception for automated driving.
\end{IEEEbiography}
\vspace{-5mm}
\begin{IEEEbiography}[{\includegraphics[width=1in,height=1.25in,clip,keepaspectratio]{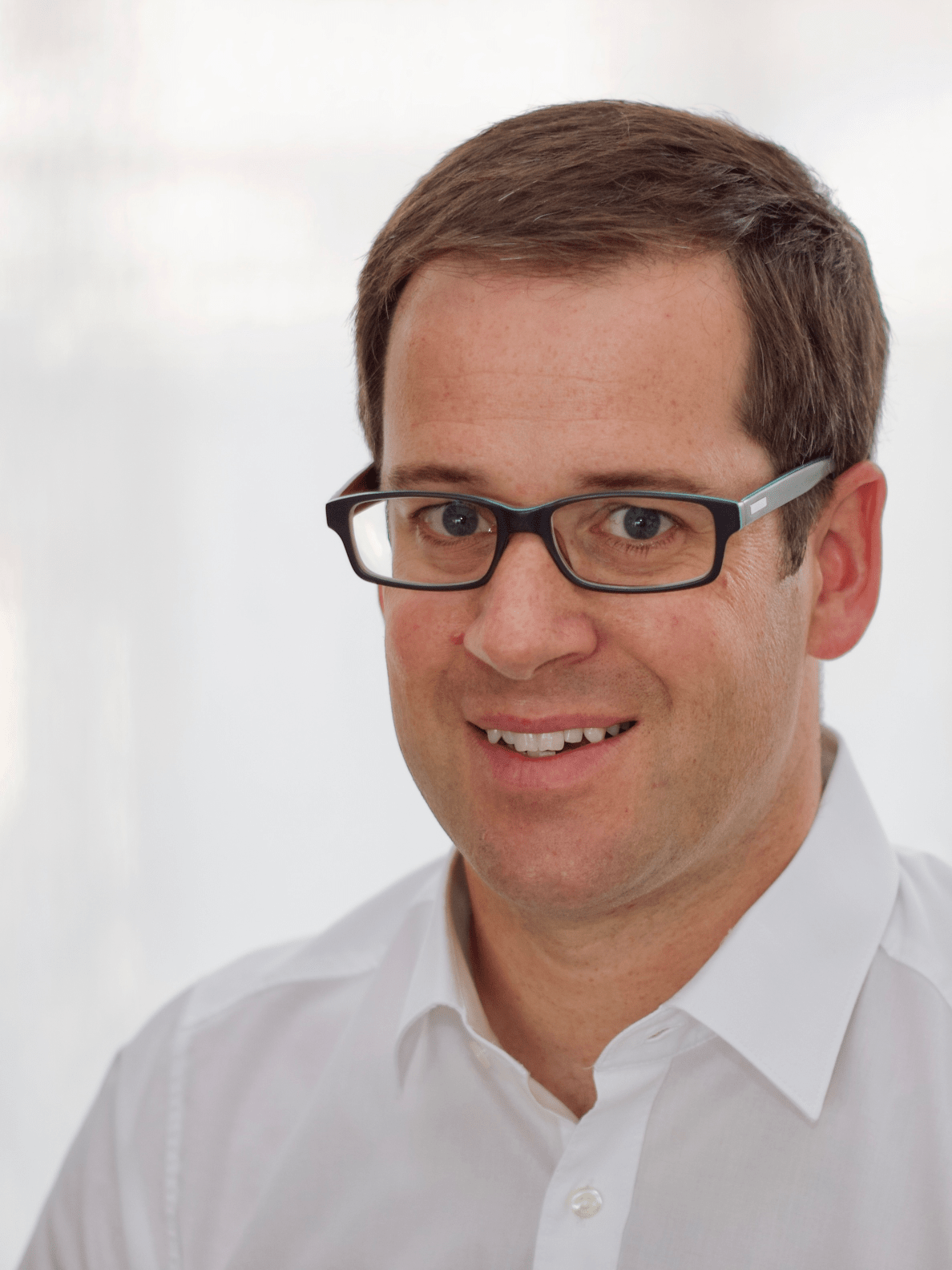}}]{Fabian Timm} studied Computer Science at the University of L\"ubeck, Germany. In 2006 he did his diploma thesis at Philips Lighting Systems in Aachen, Germany. Afterwards he started his PhD also at Philips Lighting Systems in the field of Machine Vision and Machine Learning and finished it in 2011 at the University of L\"ubeck, Institute for Neuro- and Bioinformatics. In 2012 he joined corporate research at Robert Bosch GmbH and worked on industrial image processing and machine learning. Afterwards he worked in the business unit at Bosch and developed new perception algorithms, such as pedestrian and cyclist protection only with a single radar sensor. Since 2018 he leads the research group "automated driving – perception and sensors" at Bosch corporate research. His main research interests are machine and deep learning, signal processing, and sensors for automated driving.
\end{IEEEbiography}
\vspace{-5mm}
\begin{IEEEbiography}
[{\includegraphics[width=1in,height=1.25in,clip,keepaspectratio]{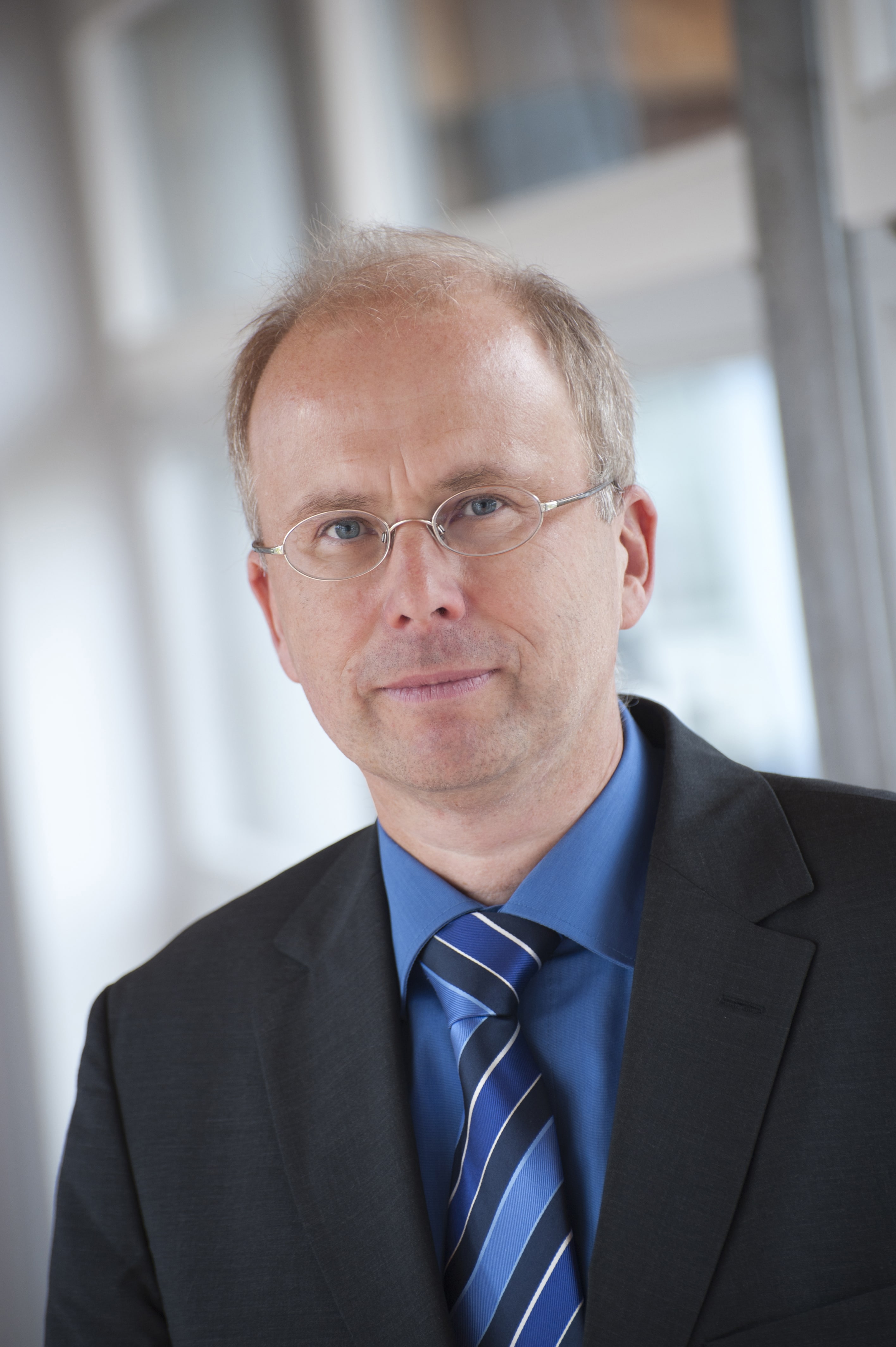}}]{Klaus Dietmayer} (Member, IEEE) was born in Celle, Germany in 1962. He received his Diploma degree in 1989 in Electrical Engineering from the Technical University of Braunschweig (Germany), and the Dr.-Ing. degree (equivalent to PhD) in 1994 from the University of Armed Forces in Hamburg (Germany). In 1994 he joined the Philips Semiconductors Systems Laboratory in Hamburg, Germany as a research engineer. Since 1996 he became a manager in the field of networks and sensors for automotive applications. In 2000 he was appointed to a professorship at the University of Ulm in the field of measurement and control. Currently he is Full Professor and Director of the Institute of Measurement, Control and Microtechnology in the school of Engineering and Computer Science at the University of Ulm. Research interests include information fusion, multi-object tracking, environment perception, situation understanding and trajectory planning for autonomous driving. Klaus Dietmayer is member of the IEEE and the German society of engineers VDI / VDE.
\end{IEEEbiography}
\vspace{-5mm}
\begin{IEEEbiography}[{\includegraphics[width=1in,height=1.25in,clip,keepaspectratio]{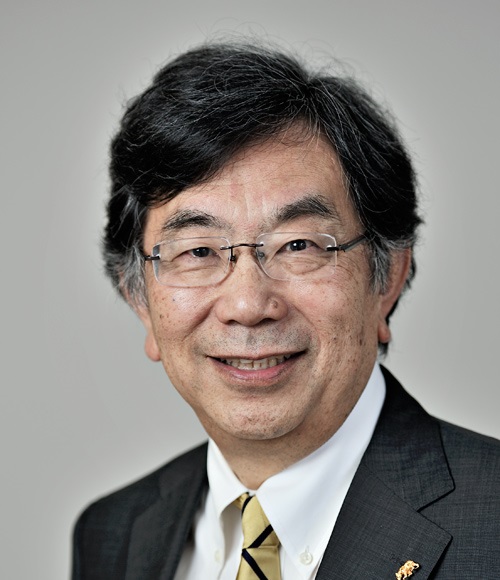}}]{Masayoshi Tomizuka} (M’86-SM’95-F’97-LF’17) received his Ph. D. degree in Mechanical Engineering from MIT in February 1974. In 1974, he joined the faculty of the Department of Mechanical Engineering at the University of California at Berkeley, where he currently holds the Cheryl and John Neerhout, Jr., Distinguished Professorship Chair. His current research interests are optimal and adaptive control, digital control, signal processing, motion control, and control problems related to robotics, precision motion control and vehicles. He served as Program Director of the Dynamic Systems and Control Program of the Civil and Mechanical Systems Division of NSF (2002- 2004). He served as Technical Editor of the ASME Journal of Dynamic Systems, Measurement and Control, J-DSMC (1988-93), and Editor-in-Chief of the IEEE/ASME Transactions on Mechatronics (1997-99). Prof. Tomizuka is a Fellow of the ASME, IEEE and IFAC. He is the recipient of the Charles Russ Richards Memorial Award (ASME, 1997), the Rufus Oldenburger Medal (ASME, 2002) and the John R. Ragazzini Award (2006).
\end{IEEEbiography}
\vspace{-5mm}
\begin{IEEEbiography}[{\includegraphics[width=1.0in,height=1.298in, clip]{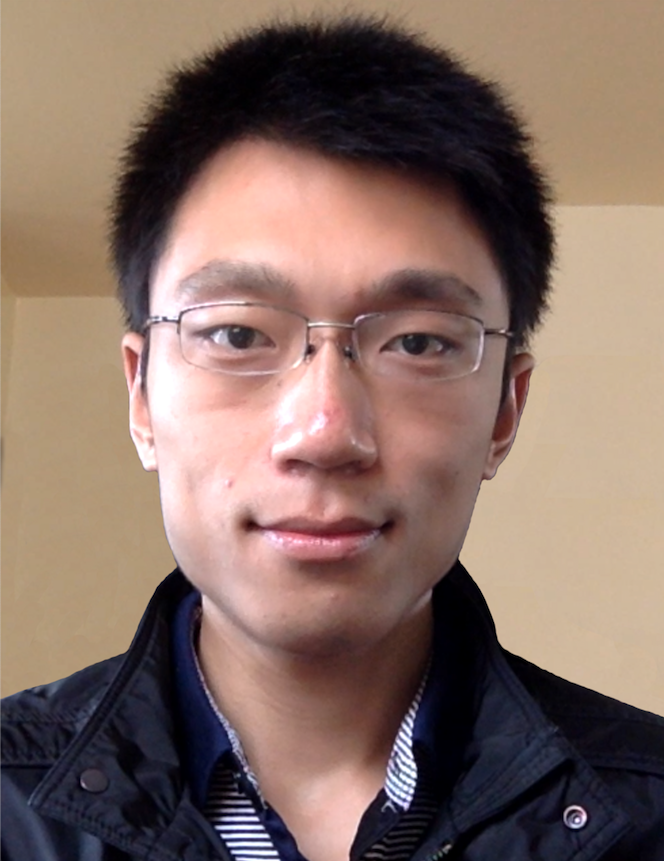}}]{Wei Zhan}
received his Ph.D. degree from University of California, Berkeley in 2019. He is currently a Postdoctoral Scholar in Mechanical Systems Control Laboratory at UC Berkeley. His main research interests lie in 3D perception, interactive behavior prediction and planning for autonomous driving with combinations of learning and model-based methods. He served as an Associate Editor in IEEE IV 2019 and IV 2020. He also served as the Chair of several workshops on behavior prediction and decision held in IEEE IV 2019, IEEE/RSJ IROS 2019, and IEEE IV 2020. He is also the lead author of the INTERACTION dataset and key organizer of the NeurIPS 2020 prediction challenge based on the dataset. One of his publications on behavior prediction for autonomous driving received Best Student Paper Award in IEEE IV 2018.
\end{IEEEbiography}
\clearpage
\newpage
\appendix
\begin{figure*}[!tpb]
	\centering
	\begin{minipage}{1\linewidth}
    \centering
	\subfigure[]{\label{fig:pscale_000032_8}\includegraphics[width=0.48\linewidth]{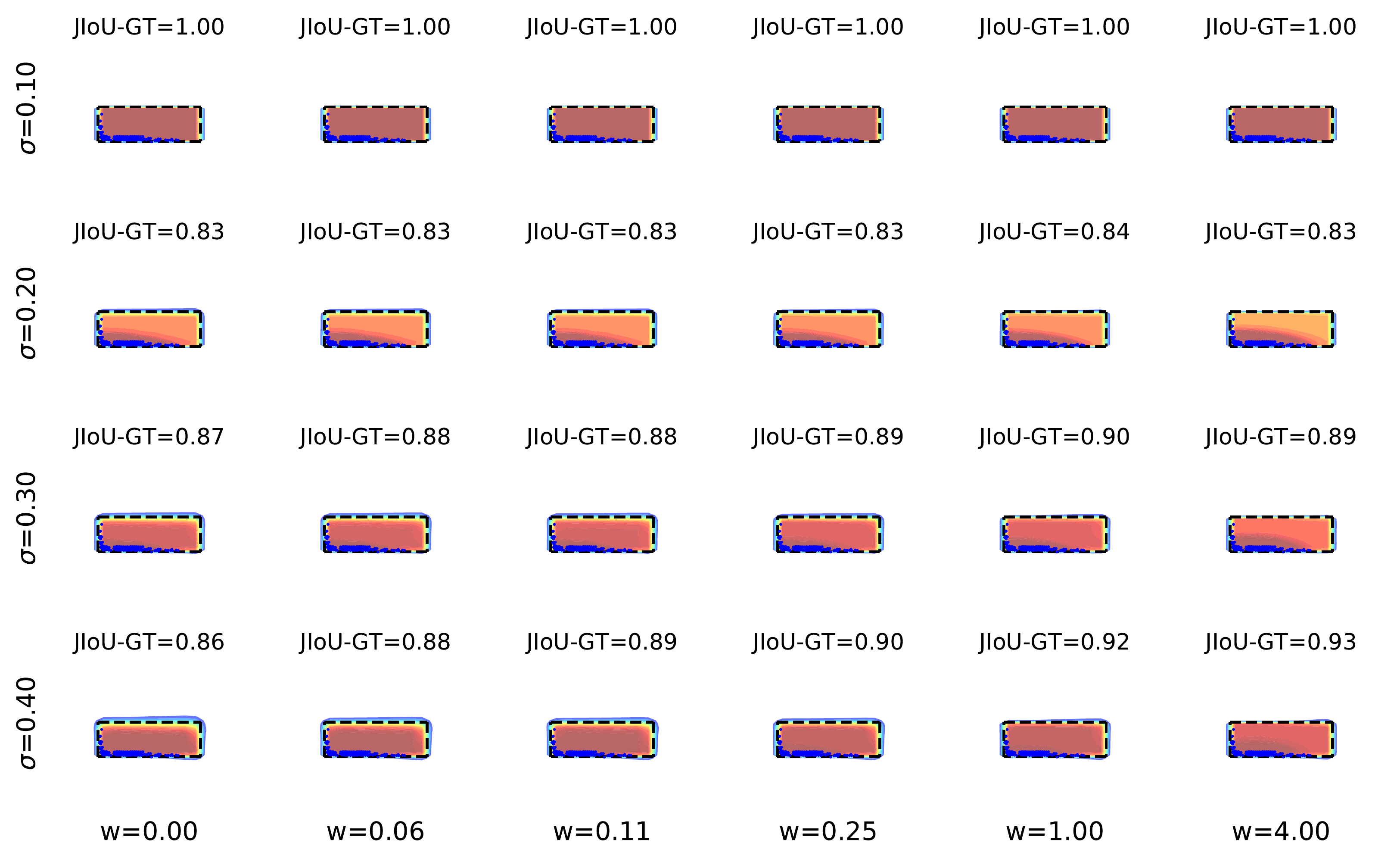}}
	\hspace{2mm}
    \subfigure[]{\label{fig:pscale_000038_2}\includegraphics[width=0.48\linewidth]{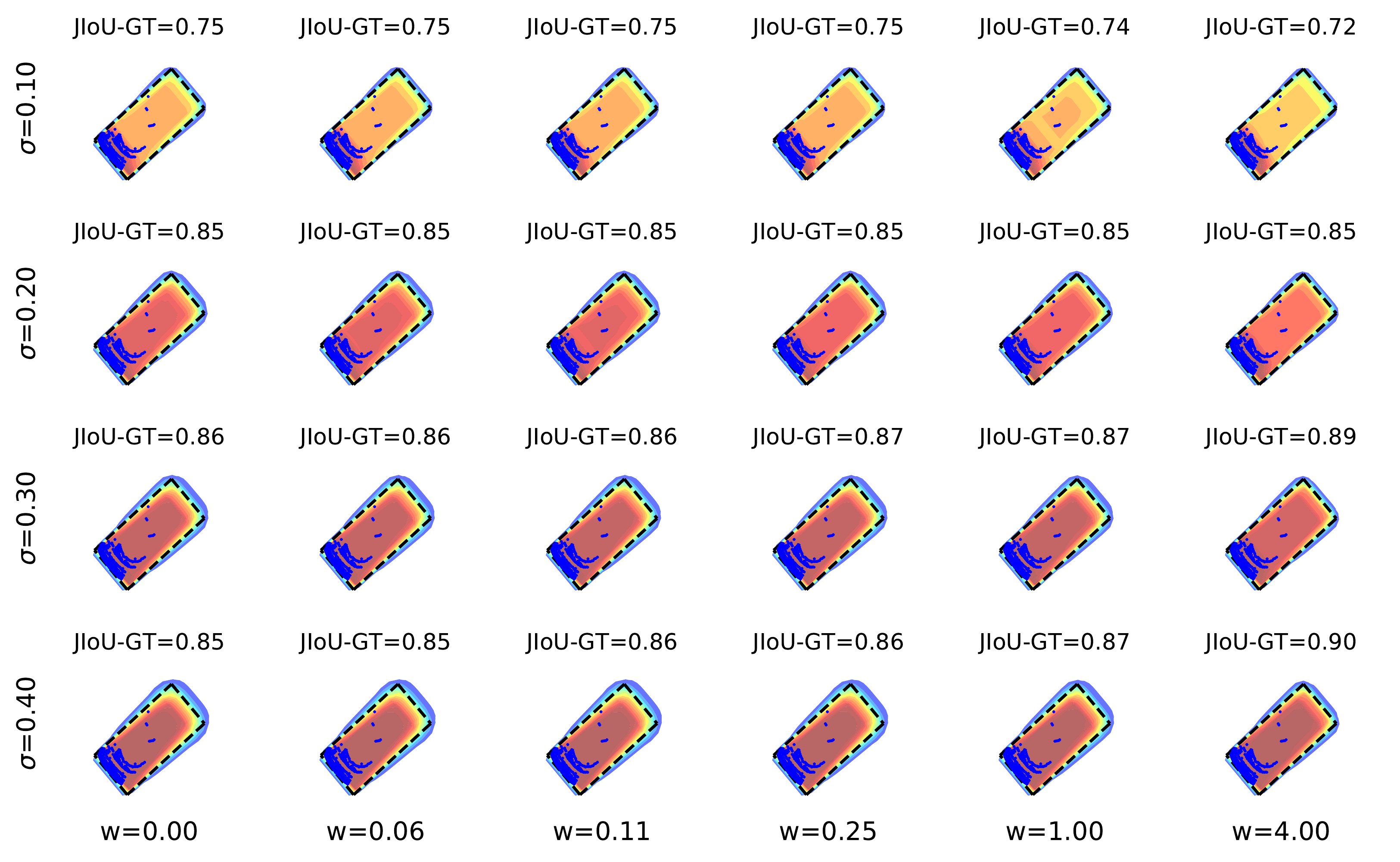}}
    \subfigure[]{\label{fig:pscale_000021_4}\includegraphics[width=0.48\linewidth]{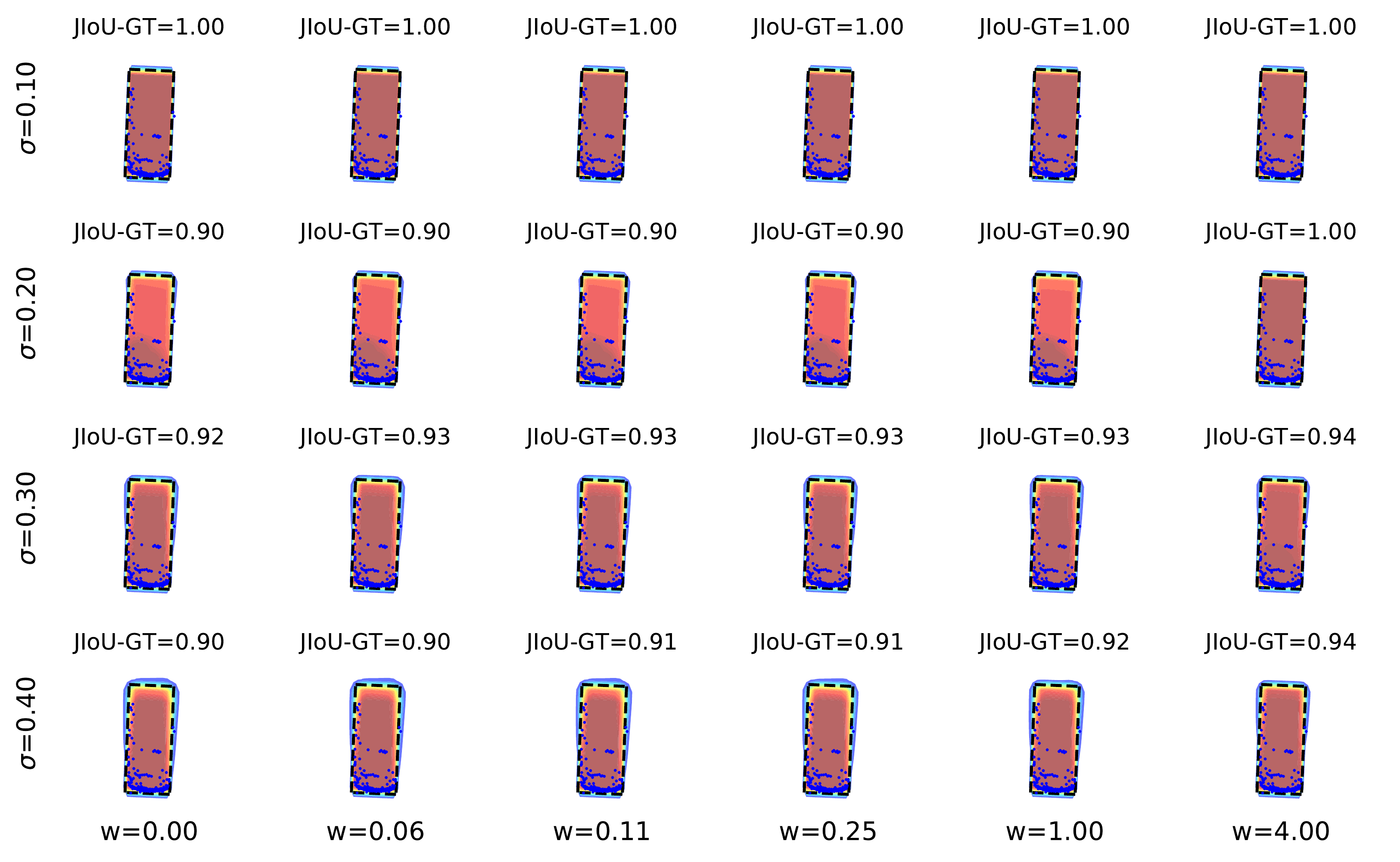}}
    \hspace{2mm}
    \subfigure[]{\label{fig:pscale_000186_0}\includegraphics[width=0.48\linewidth]{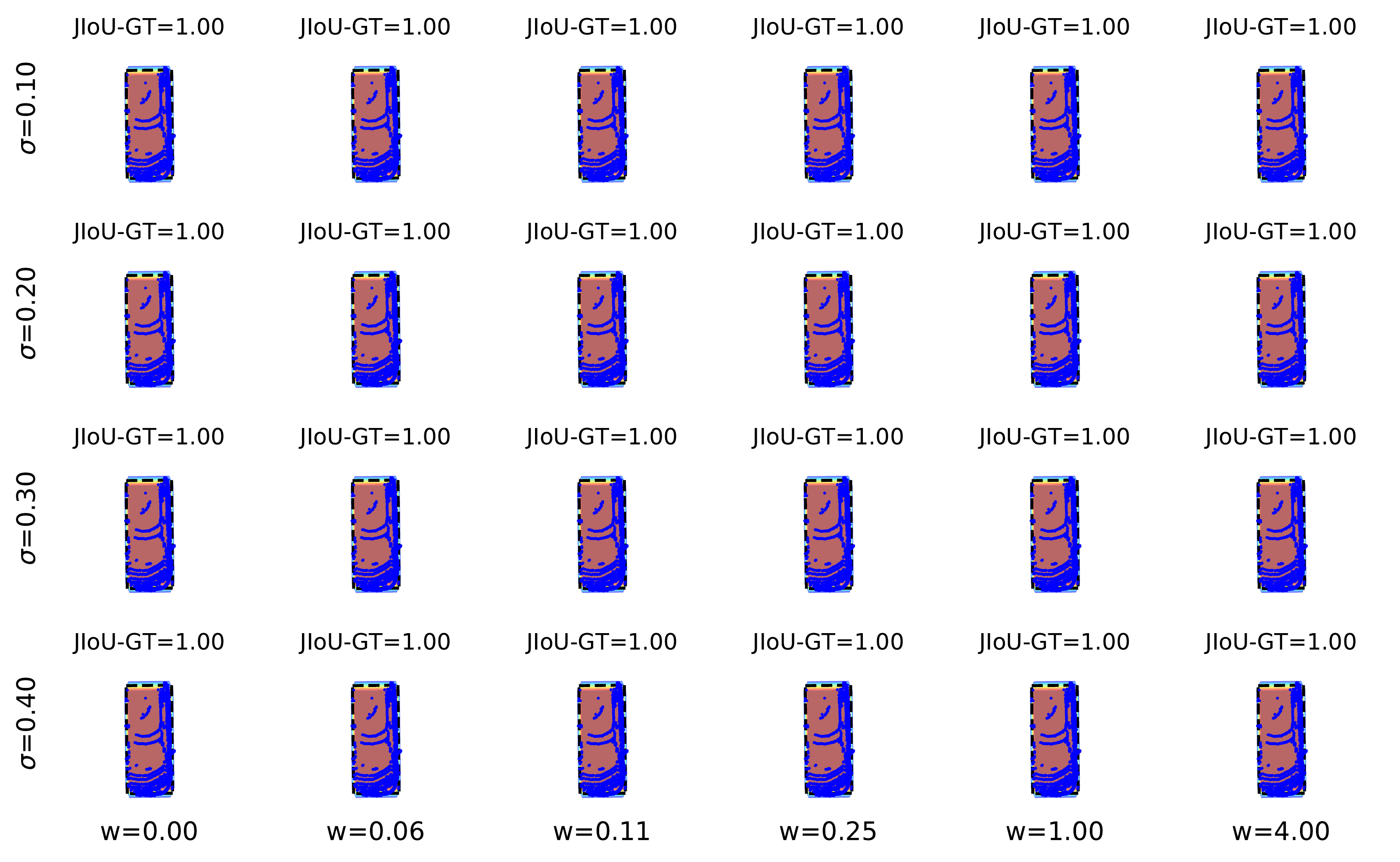}}
	\end{minipage}
	\caption{Four examples showing the impact of LiDAR observation noise $\sigma^2$ and the weight of prior knowledge $w$ on the spatial uncertainty distributions.} \label{fig:exp_parameter_influence_appendix}
\end{figure*}
\subsection{\textbf{More Experimental Results}}
\subsubsection{Choice of Parameters for the Label Uncertainty}
In Sec.~\ref{sec:uncertainty_justification:choice_of_parameters}, we have discussed the impact of LiDAR observation noise $\sigma^2$ and the weight of prior knowledge $w$ on the spatial uncertainty distributions. In this section, we present more examples. In general, the behaviours of uncertainty distributions from $\sigma$ and $w$ in Fig.~\ref{fig:exp_parameter_influence_appendix} are similar to that in Fig.~\ref{fig:exp_parameter_influence} in the main paper. However, it is observed that the label uncertainty for objects with denser LiDAR observations are less affected by the choice of parameters, such as Fig.~\ref{fig:pscale_000186_0} and Fig.~\ref{fig:pscale_000021_4}. Their distributions are quasi-uniform, indicating that our proposed label uncertainty is sensitive to the point density. 

\subsubsection{Data Distribution in the Waymo and KITTI Datasets} 
\begin{figure}[!tpb]
	\centering
	\begin{minipage}{1\linewidth}
    \centering
	\includegraphics[width=1\linewidth]{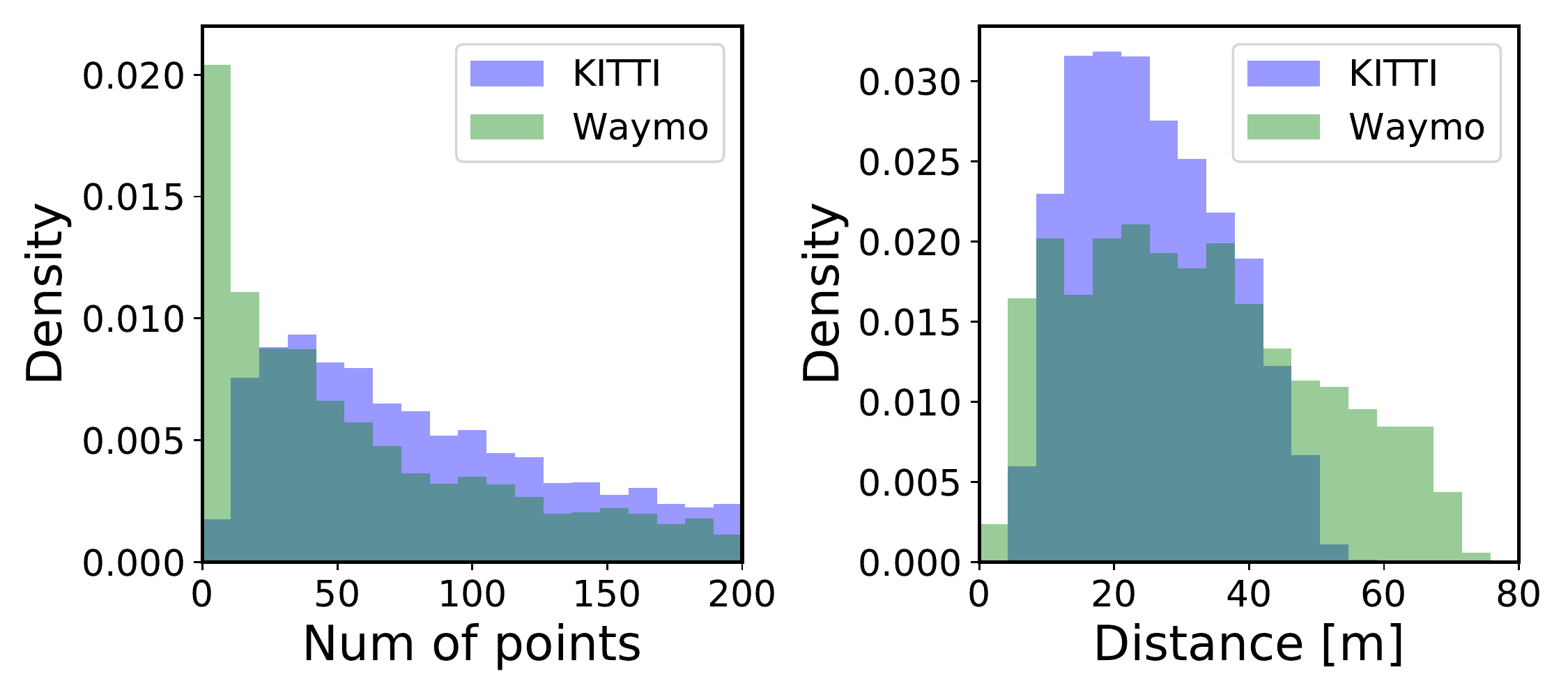}
	\end{minipage}
	\caption{Data distribution in the Waymo and KITTI datasets.} \label{fig:kitti_waymo_data_distribution}
\end{figure}

In Sec.~\ref{sec:uncertainty_justification:lidar_observation}, we have compared the label uncertainty in each corner of BBoxes in Fig.~\ref{fig:spatial_uncertainty}(a), using the total variance (TV) scores. The figure shows that the TV scores in the Waymo dataset are in general higher than those in the KITTI dataset. This is because the Waymo dataset has much more objects with sparse LiDAR points and long distance compared to KITTI, as depicted in Fig.~\ref{fig:kitti_waymo_data_distribution}. Specifically, there exist much more objects in the Waymo Dataset, whose number of LiDAR reflections are smaller than $10$ or detection distance larger than $50$ meters.


\subsection{\textbf{Proof of Spatial Uncertainty Distribution}}\label{appendix:proof_spatial_uncertainty}
This section give the equality of Eq.~\ref{eq:Generative_object} based on the relationship between $v$ and $y$ given in Eq.~\ref{eq:zy_render}. 
\begin{proof}
Denote $Y=[C_1,C_2,C_3,L,W,H,r_Y]^T$, $C{:=}[C_1,C_2,C_3]^T$, $S{:=}[L,W,H]$, where the capital letters indicate random variables. Also denote $R_Y$ as the rotation matrix of $r_Y$ and $v^*{:=}[v_1,v_2,v_3]^T$. From Eq.~\ref{eq:zy_render}, we have
\begin{equation}
     R_Y^T(V{-}C)=\left[\begin{smallmatrix}l\\w\\h\end{smallmatrix}\right]\circ v^*,
\end{equation}
where $\circ$ denotes the element-wise multiplication (Hadmard product) of matrices. This equation means when conditioned on $r_Y{=}r_y$ and $C=[c_1,c_2,c_3]$, the mapping between $V$ and $S$ is a diffeomorphism and the probability density function of $V|r_Y,C$ is
\begin{equation}
    p_{V|r_y,c}(u) = \left|\frac{1}{v_1v_2v_3}\right|p_{S|r_y,c}\left
    (R_y^T(u{-}c)\circ v^{*^{-1}}\right),
\end{equation}
where $v^{*^{-1}}$ is the element-wise inverse of $v^*$. Then
\begin{equation}
    \begin{aligned}
    &\quad\ \int_{v^*\in B(y^*)}p_{V}\left(u\right)dv^* \\
    &=\int_{v^*\in B(y^*)}\left(\int{p_{V|r_y,c}(u)}p_{r_Y,C}(r_y,c)dr_ydc\right)dv^*\\
    &=\int{p_{r_Y,C}dr_ydc\int_{v^*\in B(y^*)}{\left|\frac{1}{v_1v_2v_3}\right|p_{S|r_y,c}\left
    (R_y^T(u{-}c)\circ v^{*^{-1}}\right)dv^*}}\\
    &=\int{p_{r_Y,C}dr_ydc \int_{\{s|u\in B(c,s,r_y)\}}{\left|\frac{1}{l w h}\right|p_{S|r_y,c}\left
    (s\right)dldwdh}}\\
    &=\int_{\{c,s,r_y|u\in B(y),y=[c,s,r_y]\}}{\left|\frac{1}{l w h}\right|p_Y(c,s,r_y)dc ds dr_y}
    \end{aligned}.
\end{equation}
where $s{=}[l,w,h]^T$ are the length, width and height of the BBox. Note that $\|lwh\|=A(y)$, hence Eq.~\ref{eq:JIoU_def} is proved.
\end{proof}

\subsection{\textbf{How to Calculate Spatial Uncertainty Distribution}}\label{appendix:howto_spatial_uncertainty}
The distribution $p_{V(v^*,Y)}(u)$ of $V(v^*,Y)$ is required for the calculation of the spatial uncertainty distribution defined by Eq.~\ref{eq:Generative_object}. Several possible situations may occur:
\begin{enumerate}
    \item The distributions of $V(v^*,Y)$ is derived through Eq.~\ref{eq:zy_render_linear_form} and Eq.~\ref{eq:VB_solution_covariance} in the label uncertainty inference method.
    \item The distributions of some points of $V(v^*,Y)$, such as corners of the BBox~\cite{hall2018probabilistic,meyer2019lasernet}, are predicted.
    \item The distribution of $Y$ is predicted but the distribution of $V$ is unknown, which are the cases of~\cite{feng2018leveraging,wirges2019capturing} and Section~\ref{sec:uncertainty}
\end{enumerate}

In the case of 1), it is easy to calculate $p_G(u)$ by doing a numerical integration on $p_V(v)$. In the case of 2) and 3), we need to calculate the distribution of $V(v^*,Y)$ first. The moments of $V(v^*,Y)$ is related to $Y$ by Eq.~\ref{eq:zy_render}, which can be rewritten in the homogeneous coordinate of $v^*$ as
\begin{equation}
    v(v^*,y)=\Phi(y)^T\xi,
\end{equation}
where $\xi$ is the homogeneous coordinate of $v^*$ and 
\begin{equation}
    \Phi(y)^T=\begin{bmatrix}
                l\cos(r_y) & 0 & -w\sin(r_y) & c_1\\
                0 & h & 0 & c_2\\
                l\sin(r_y)& 0 & w\cos(r_y) & c_3
                \end{bmatrix},
\end{equation}
is the feature matrix. Considering the assumption that we approximate the distribution of $V(v^*,Y)$ as Gaussian, we only need to calculate 
\begin{equation}
    \begin{aligned}
    E[V] &= E[\Phi(Y)]^T\xi\\
    E[V^2] &= E[\Phi(Y)^T\xi\xi^T\Phi(Y)],
    \end{aligned}
\end{equation}
for all $\xi$'s corresponding to the points $v^*$ in the \textit{unit} BBox. For 3D, $\xi\xi^T\in \mathbb{R}^4$ is a symmetric matrix. Since the base of the set $Sym_4\left(\mathbb{R}\right)$ of symmetric matrices is 10, at most 10 points is enough for representing all possible $\xi\xi^T$'s. This means there exists a set $\xi_1,\xi_2,\cdots, \xi_{10}$ such that for any homogeneous coordinate $\xi$, there exist coefficients $\alpha_1, \alpha_2, \cdots, \alpha_{10}$ such that
\begin{equation}
    \sum_{i=1}^{10}{\alpha_i \xi_i\xi_i^T} = \xi\xi^T.
\end{equation}

Note that $\xi$ is homogeneous coordinate which means the 4th colum of $\xi\xi^T$ is $\xi$, so there is also
\begin{equation}
    \sum_{i=1}^{10}{\alpha_i \xi_i} = \xi.
\end{equation}

When we want to recover the uncertainty of parameters, specific $\xi$'s can be chosen. For the center $(c_1,c_2,c_3)$, choose $\xi_c=[0,0,0,1]^T$ so that
\begin{equation}
    E[\Phi(Y)\xi_c\xi_c^T\Phi(Y)]=E\left[\begin{bmatrix}
                c_1^2 & 0 & 0\\
                0 & c_2^2 & 0\\
                0& 0 & c_3^2
                \end{bmatrix}\right],
\end{equation}
gives the variance of the center parameters. For the length $l$, choose $\xi_l=[1,0,0,0]^T$, so that
\begin{equation}
\begin{aligned}
    E[l^2] &= trace\left(E[\Phi(Y)\xi_l\xi_l^T\Phi(Y)]\right)\\
    &=trace\left(E\left[\begin{bmatrix}
                l^2\cos^2(r_y) & 0 & 0\\
                0 & 0 & 0\\
                0& 0 & l^2\sin^2(r_y)
                \end{bmatrix}\right]\right),
\end{aligned}
\end{equation}

and for the width $\xi$, choose $\xi_w=[0,0,1,0]^T$ so that
\begin{equation}
\begin{aligned}
    E[w^2] &= trace\left(E[\Phi(Y)\xi_w\xi_w^T\Phi(Y)]\right)\\
    &=trace\left(E\left[\begin{bmatrix}
                w^2\sin^2(r_y) & 0 & 0\\
                0 & 0 & 0\\
                0& 0 & w^2\cos^2(r_y)
                \end{bmatrix}\right]\right),
\end{aligned}
\end{equation}
and for the height $h$, choose $\xi_h=[0,1,0,0]^T$ so that
\begin{equation}
\begin{aligned}
    E[h^2] &= trace\left(E[\Phi(Y)\xi_h\xi_h^T\Phi(Y)]\right)\\
    &=trace\left(E\left[\begin{bmatrix}
                0 & 0 & 0\\
                0 & h^2 & 0\\
                0& 0 & 0
                \end{bmatrix}\right]\right),
\end{aligned}
\end{equation}

\end{document}